\documentclass[acmtog, authorversion]{acmart}

\usepackage{amsmath, amsfonts, amsthm, bm}
\usepackage{wrapfig}
\usepackage{colortbl}
\usepackage[normalem]{ulem}

\AtBeginDocument{%
  \providecommand\BibTeX{{%
    \normalfont B\kern-0.5em{\scshape i\kern-0.25em b}\kern-0.8em\TeX}}}

\setcopyright{cc}
\setcctype{by}
\acmJournal{TOG}
\acmYear{2025} \acmVolume{44} \acmNumber{4} \acmArticle{} \acmMonth{8} \acmPrice{}\acmDOI{10.1145/3730922}

\begin{document}
\title{Neural Co-Optimization of Structural Topology, Manufacturable Layers, and Path Orientations for Fiber-Reinforced Composites}

\author{Tao Liu}\orcid{0000-0003-1016-4191}
\authornotemark[1]
\affiliation{
  \institution{The University of Manchester}
  \city{Manchester}
  \country{United Kingdom}
}

\author{Tianyu Zhang}\orcid{0000-0003-0372-0049}
\authornote{Joint first authors.}
\affiliation{
  \institution{The University of Manchester}
  \city{Manchester}
  \country{United Kingdom}
}

\author{Yongxue Chen}\orcid{0000-0001-6236-4158}
\affiliation{
  \institution{The University of Manchester}
  \city{Manchester}
  \country{United Kingdom}
}

\author{Weiming Wang}\orcid{0000-0001-6289-0094}
\affiliation{
  \institution{The University of Manchester}
  \city{Manchester}
  \country{United Kingdom}
}

\author{Yu Jiang}\orcid{0000-0003-1579-7158}
\affiliation{
  \institution{The University of Manchester}
  \city{Manchester}
  \country{United Kingdom}
}

\author{Yuming Huang}\orcid{0000-0001-5900-2164}
\affiliation{
  \institution{The University of Manchester}
  \city{Manchester}
  \country{United Kingdom}
}

\author{Charlie C. L. Wang}\orcid{0000-0003-4406-8480}
\authornote {Corresponding author: changling.wang@manchester.ac.uk (Charlie C.L. Wang).  }
\affiliation{
  \institution{The University of Manchester}
  \city{Manchester}
  \country{United Kingdom}
}

\begin{abstract}
We propose a neural network-based computational framework for the simultaneous optimization of structural topology, curved layers, and path orientations to achieve strong anisotropic strength in fiber-reinforced thermoplastic composites while ensuring manufacturability. Our framework employs three implicit neural fields to represent geometric shape, layer sequence, and fiber orientation. This enables the direct formulation of both design and manufacturability objectives -- such as anisotropic strength, structural volume, machine motion control, layer curvature, and layer thickness -- into an integrated and differentiable optimization process. By incorporating these objectives as loss functions, the framework ensures that the resultant composites exhibit optimized mechanical strength while remaining its manufacturability for filament-based multi-axis 3D printing across diverse hardware platforms. Physical experiments demonstrate that the composites generated by our co-optimization method can achieve an improvement of up to $33.1\%$ in failure loads compared to composites with sequentially optimized structures and manufacturing sequences.
\end{abstract}

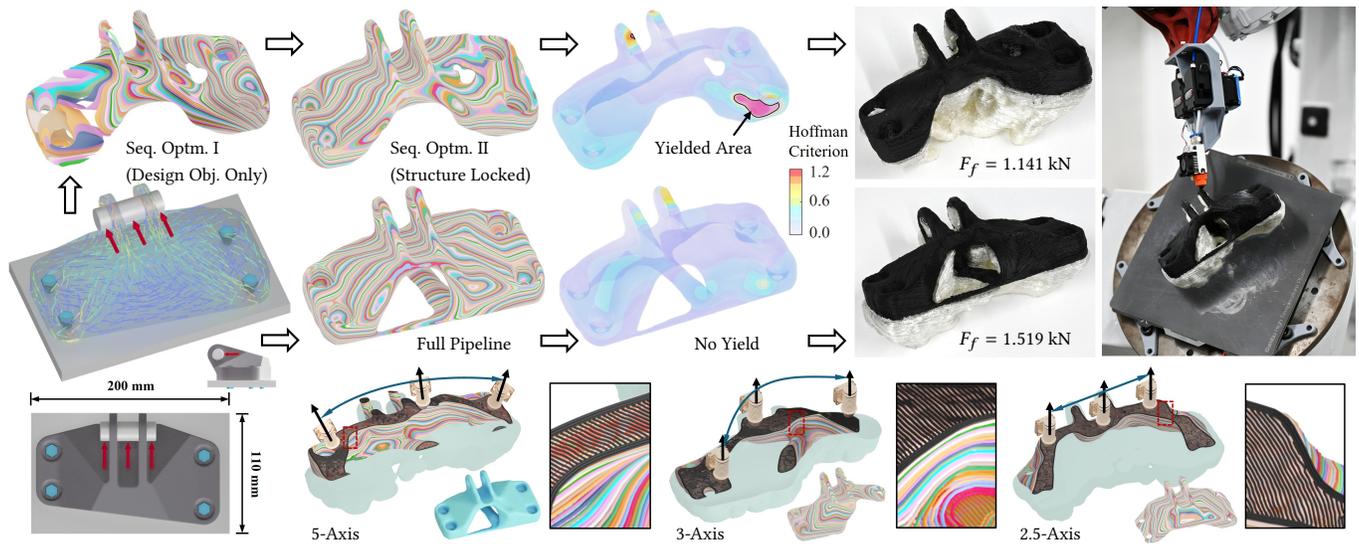
\begin{teaserfigure}
\centering
\includegraphics[width=\textwidth]{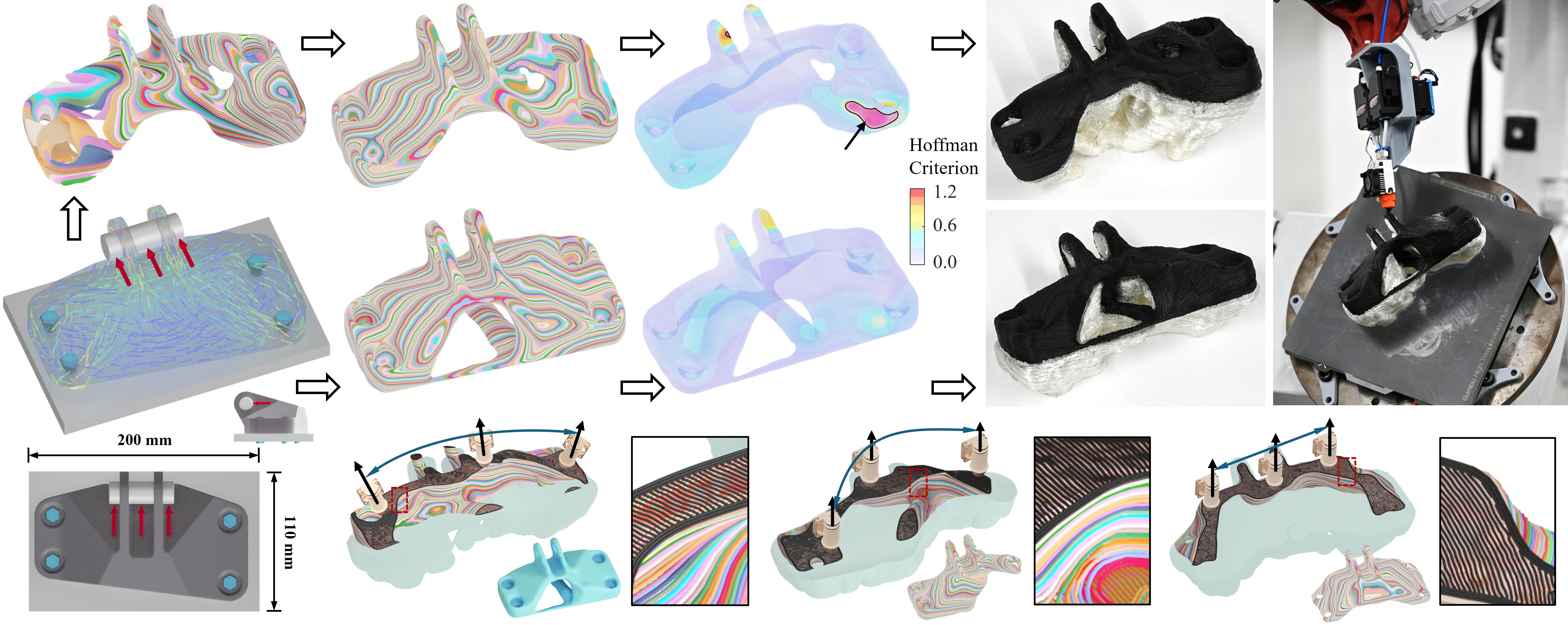}
\put(-465,148){\footnotesize \color{black}Seq. Optm. I}
\put(-465,138){\footnotesize \color{black}(Design Obj. Only)}
\put(-365,148){\footnotesize \color{black}Seq. Optm. II}
\put(-365,138){\footnotesize \color{black}(Structure Locked)}
\put(-355,74){\footnotesize \color{black}Full Pipeline}
\put(-395,2){\footnotesize \color{black}5-Axis}
\put(-257,2){\footnotesize \color{black}3-Axis}
\put(-127,2){\footnotesize \color{black}2.5-Axis}
\put(-265,148){\footnotesize \color{black}Yielded Area}
\put(-250,74){\footnotesize \color{black}No Yield}
\put(-150,142){\footnotesize \color{black}$F_f = 1.141~\text{kN}$}
\put(-150,76){\footnotesize \color{black}$F_f = 1.519~\text{kN}$}
\caption{
Compared to the fiber-reinforced composite optimized using our method that simultaneously optimizes the topological structure, corresponding layers, and path orientations, the result from sequential design and manufacturing optimization exhibits significantly larger yielded areas. These areas are indicated by regions with a failure index value exceeding 1.0, based on the Hoffman Criterion~\cite{Schellekens1990}. This observation is further validated through mechanical tests, where failure loads $F_f$ were measured on specimens fabricated by a multi-axis 3D printing system. Our co-optimization framework incorporates direct formulation of manufacturing constraints and is able to support 3D printing hardware with varying degrees-of-freedom (i.e., 5-axis, 3-axis, and 2.5-axis motions). Note that a target volume fraction $15\%$ is used for this GE-Bracket challenge problem. 
}\label{fig:teaser}
\vspace{5pt}
\end{teaserfigure}



\maketitle
\section{Introduction}\label{sec:Intro}
Fiber-reinforced composites, with their exceptional properties of high stiffness, high strength, and low density in carefully designed parts, have gained increasing attention in engineering design and manufacturing as key solutions for creating lightweight structures with superior mechanical performance ~\cite{CHENG2023CFRPSurvey,GONZALEZ2017194}. Multi-axis 3D printing techniques that allow dynamic control over material deposition orientations have emerged as an excellent means for fabricating such composites. However, prior research has largely employed two-stage approaches -- first optimizing the structural topology of a part and then computing curved layers and fiber toolpaths to align with principal stress directions (e.g., \cite{fang2020reinforced,tam20173d}). Unlike topology optimization for isotropic materials ~\cite{Sigmund1994PHDTHESIS,Sigmund2001120}, recent efforts have begun to optimize fiber orientations within the composite volume alongside the structural topology (e.g.,~\cite{Luo2023Spatially,QIU2022Concurrent}). Concurrent optimization methods have also been proposed to jointly compute topology and manufacturing sequences for robotized 3D printing ~\cite{Wang20201}. However, a common limitation of these existing approaches is the indirect formulation of manufacturability constraints -- such as machine motion \textit{degrees-of-freedom} (DoFs), layer curvatures, and path curvatures -- which are either inadequately addressed or entirely overlooked.

When manufacturing constraints are not properly incorporated during the topology optimization phase, the mechanical strength of the finally fabricated parts can be compromised. Due to manufacturability requirements, the layers and toolpaths generated in a subsequent step may result in fiber orientations that fail to satisfy the Hoffman criterion for yield. As shown in Fig.\ref{fig:teaser}, optimizing solely based on design objectives -- i.e., structural topology and fiber orientations -- leads to layers with large curvatures (which may cause collisions between layers and the printer head) and significant thickness variations (which will be a problem for fiber filament deposition). Addressing these issues in a subsequent manufacturing optimization step with unchanged structural topology results in a model with large areas to yield. The yielding analysis shown in this figure was performed using a commercial \textit{Finite Element Analysis} (FEA) software, Abaqus. These simulation results have been validated by mechanical tests, where the failure loads $F_f$ are measured on the specimens fabricated by a multi-axis 3D printing system. The failure load is improved from $F_f=1.141\text{kN}$ to $F_f=1.519\text{kN}$ on the model generated by co-optimization. Moreover, incorporating constraints with varying machine motion DoFs into the co-optimization process leads to distinct optimal structures for fiber-reinforced composites (see the bottom of Fig.\ref{fig:teaser}).

\subsection{Method and Contribution}
In this paper, we present a neural network-based inverse design framework that simultaneously optimizes the topological structure and the manufacturing sequence together with fiber orientations for fiber-reinforced composites. The yield failure of designed composites with strong anisotropic strength is evaluated by the Hoffman criterion (Sec.~\ref{subsec:HoffmanCriterion}). The manufacturability of these composites via multi-axis 3D printing is ensured by directly formulating major manufacturing constraints -- such as machine motion DoFs, layer curvatures, path curvatures, and layer thickness -- into differentiable loss functions. Our framework conducts three neural implicit fields to provide design and manufacturing information at any query point $\mathbf{x} \in \mathbb{R}^3$ within the design domain (Sec.~\ref{subsec:NeuralFieldRep}), including: 
\begin{enumerate}
    \item a scalar density field $\rho(\mathbf{x},\theta_\rho)$ for material distribution, 

    \item a scalar field $m(\mathbf{x},\theta_m)$ for the order of material deposition (i.e., the layers of 3D printing);

    \item an auxiliary scalar field $a(\mathbf{x},\theta_a)$ for helping define the fiber orientation field together with $m(\mathbf{x},\theta_m)$.
\end{enumerate}
Facilitated by this field-based representation, the material coordinate frames are well defined to conduct the anisotropic FEA (Sec.~\ref{subsec:AnisotropicFEA}) for evaluating the mechanical strength by the Hoffman criterion. The neural network parameters $\theta_\rho$, $\theta_m$ and $\theta_a$ serve as the design variables to be optimized via FEA-integrated backpropagation (Sec.~\ref{subsec:CompPipeline}) based on the differentiable loss functions (Sec.~\ref{sec:LossFunc}), evaluating both mechanical performance and manufacturability.

Our work introduces the following technical contributions. 
\begin{itemize}
\item We propose a computational framework that employs implicit neural representation to simultaneously optimize design and manufacturing objectives, outperforming sequential optimization methods that often compromise mechanical strength to meet manufacturability. 

\item We directly formulate manufacturing constraints of multi-axis 3D printing as loss functions in the optimization process, ensuring that the resultant layers and path orientations are manufacturable across diverse hardware systems.

\item We develop strength-based topology optimization formulas based on the Hoffman criterion, specifically tailored for fiber-reinforced composites that exhibit highly anisotropic strength and can be fabricated by multi-axis 3D printing. 
\end{itemize}
As a general framework, our computational pipeline accommodates hardware with different DoFs in motion, providing practical solutions for advanced 3D printing applications. To validate the effectiveness of our approach, we have conducted comprehensive numerical simulations and physical experiments on a variety of examples.

\begin{table*}[t]
    \caption{Comparison of most relevant methods in literature}
\vspace{-5pt}
\centering
    \label{tab:methodComparison}
\footnotesize
    \centering
    \begin{tabular}{l|l||c|c|c|c|c}
        \hline
         &   & \multicolumn{5}{c}{Design \& Manufacturing Objectives} \\
       \cline{3-7}  
       Methods  & Representation & (A) Strength &  (B) Fiber-Orientation &  (C) Layer-Thickness & (D) Tool-Collision  &  (E) Motion DoFs \\
       \hline \hline
       \cite{Wang20201}  & \textit{Two} 3D Scalar Fields  & \cellcolor{red!25}No   & \cellcolor{red!25}No   & \cellcolor{green!25}Yes   & \cellcolor{red!25}No   & \cellcolor{red!25}No \\
       \cite{Luo2023Spatially} & \textit{Four} 3D Scalar Fields &  \cellcolor{red!25}No  &  \cellcolor{green!25}Yes  &  \cellcolor{red!25}No  &  \cellcolor{red!25}No  &  \cellcolor{red!25}No \\
       \cite{Li2024Strength} & \textit{One} 2D Scalar Field + \textit{One} 2D Vector Field &  \cellcolor{green!25}Yes  &  \cellcolor{green!25}Yes  &  \cellcolor{red!25}No  &  \cellcolor{red!25}No &  \cellcolor{red!25}No \\
       Ours &  \textit{Three} 3D Scalar Fields &  \cellcolor{green!25}Yes  &  \cellcolor{green!25}Yes  &  \cellcolor{green!25}Yes  &  \cellcolor{green!25}Yes  &  \cellcolor{green!25}Yes \\
    \hline
    \end{tabular}
\end{table*}

\subsection{Related works}
\subsubsection{Multi-Axis 3D Printing}
While planar layer-based 3D printing has been widely used in different engineering applications (e.g., ~\cite{Panetta2017Worst, romain2013make}), studies such as~\cite{fang2020reinforced, Xavier2023Orientable} have highlighted its limitations, particularly its weak inter-layer mechanical strength. Incorporating additional DoFs in 3D printing offers exciting opportunities and greater flexibility to customize anisotropic mechanical strength within the volume of a fabricated model. Researches in computational fabrication have been conducted by using machines with multi-axis motions -- such as modified desktop 3D printers~\cite{etienne2019curvislicer}, 5-axis CNC machines~\cite{bartovn2021geometry, zhong2023vasco}, and robotic systems~\cite{duenser2020robocut, huang2016framefab}.

Existing approaches in computational design and fabrication often frame the tasks as a field-based optimization problem (e.g.,~\cite{montes2023differentiable, arora2019volumetric, Stutz2022Synthesis, Wu202143, XU2024117270}. Following this strategy, those methods for multi-axis 3D printing usually first compute an optimized field within the volume to generate curved layers and subsequently extend the field to the surfaces for toolpath generation. Various types of fields have been employed. For example, the scalar fields are employed for curved layers in ~\cite{dai2018support-free} and for toolpaths in ~\cite{Zhao2016Connected}. Vector fields combined with scalar fields were employed in ~\cite{fang2020reinforced} to generate curved layers and toolpaths, aligning with principal stresses. A deformation-based optimization was introduced in ~\cite{etienne2019curvislicer} to compute a height field for slightly curved layers suitable for 3-axis printing. Recently, deformation fields are optimized in ~\cite{Zhang2022s3, Liu2024NeuralSlicer} for generating curved layers in 5-axis 3D printing. However, maintaining a fixed structural topology often requires greater compromises in mechanical strength for satisfying manufacturing constraints of fibre-reinforced composites. In short, stronger composites are expected when optimizing the topological structures together with manufacturing means.

\subsubsection{Topology Optimization}
Topology optimization (TO) has become more and more widely employed in the practice of engineering design (ref.~\cite{Sigmund1994PHDTHESIS, Sigmund2009227, Nguyen2020Efficient}) to determine the material distribution within a given design domain to maximize performance. Over the years, various topology optimization algorithms have been developed, including evolutionary structural optimization~\cite{Huang200889, Huang2010671, Xia2018437}, solid isotropic material with penalization (SIMP)~\cite{Sigmund199868, Sigmund2001120, Andreassen20111}, level set methods~\cite{Sethian2000489, Wang2003}, and moving morphable components~\cite{Guo2016711}. These methods were primarily designed for isotropic materials, and therefore often fail to account for anisotropic mechanical properties. 

More recent advancements in optimization have begun addressing anisotropic properties in various applications, such as fluid flows~\cite{Li2022Fluidic}, multi-material structures~\cite{Yan2020Strong}, and microstructures~\cite{Wu202143, arora2019volumetric, JENSEN2024Efficient, Zhu2017TwoScale}. Incorporating anisotropic mechanical analysis into topology optimization is particularly beneficial for fiber-reinforced composites~\cite{Luo2023Spatially}, as it better captures their unique material behaviors.

In addition to mechanical properties, there has been a growing focus on incorporating manufacturability considerations into TO. Different aspects of manufacturing have been addressed, including the need for support structures~\cite{Wang2023, Langelaar201660}, the distortion during printing~\cite{MIKI2021Topology}, and the accessibility of closed holes~\cite{YAMADA2022Topology}. A recent trend in TO is to concurrently optimize a structure's topology together with the manufacturing sequence~\cite{Wang2023Temporal, WU2024Spacetime, WANG2024Regularization} or printing orientations~\cite{Ye2023}. Besides the support-free TO for conventional planar printing, many other TO approaches for multi-axis 3D printing lack of physical validation. This is primarily due to insufficient consideration of comprehensive manufacturability constraints -- such as motion DoFs and collision avoidance -- considered in our approach.

\subsubsection{Optimization for Fiber-Reinforced Composites}
Optimization techniques have been widely applied in the design and manufacturing of fiber-reinforced composites, where 3D printing gains significant attention in recent years. Comprehensive reviews on this topic can be found in a survey paper such as \cite{CHENG2023CFRPSurvey}, and we only discuss the most relevant works below.

In earlier approaches, researchers aligned continuous carbon fibers with the stress field to enhance the mechanical strength of composites by applying optimized toolpaths within each planar layer~\cite{CHEN2022FieldBased,LIU2023Stress}. More recently, Fang et al.~\shortcite{Fang2024AMCCF} extended their previous work~\cite{fang2020reinforced} to generate curved layers for printing fiber-reinforced composites while considering fiber continuity. To address the issue of sparse toolpaths when generating them directly from vector and scalar fields, 2-RoSy representation and periodic parameterization, akin to the method of \cite{Knoppel2015Stripe}, were employed in \cite{zhang2024toolpath} to create dense 3D printing toolpaths for continuous carbon fibers. However, without simultaneously optimizing the topological structures, the generation of curved layers and toolpaths often sacrifices mechanical strength to ensure manufacturability.

Level set functions \cite{Xu2024} and neural implicit functions \cite{Chandrasekhar2023} have been used to formulate the concurrent optimization of structural topology and fiber orientations for 2D problems. Manufacturing requirements, such as maintaining a nearly uniform distance between fiber toolpaths, are typically addressed in the post-processing stage -- i.e., outside the optimization loop. Recent advancements in \cite{Ren2024TopoFiber} have begun to incorporate these 2D manufacturing requirements directly into the optimization loop by employing a wave function. However, more comprehensive manufacturing constraints for 3D problems remain unexplored. On the other aspect, all these approaches are based on stiffness-driven optimization, which may not be well-suited for yield-related problems. Strength-based formulation for 2D problems was proposed in \cite{Li2024Strength}, which inspires our formulation of strength-based TO for 3D problems based on the Hoffman criterion. 

With advancements in modern computing techniques, such as GPU acceleration~\cite{qi2025Efficient}, researchers have been able to concurrently optimize 3D topological structures and fiber orientations~\cite{QIU2022Concurrent,Luo2023Spatially}. However, these approaches either neglect manufacturing requirements or incorporate only simplistic considerations, such as the sequence of printing orientations~\cite{GUO2025Concurrent}. No existing TO approach for fiber-reinforced composites has addressed the complex and comprehensive manufacturing constraints as proposed in this paper, including motion DoFs, collision avoidance, and layer thickness control~(see the comparison summarized in Table \ref{tab:methodComparison}).

The primary challenge in integrating existing TO approaches (e.g., \cite{Li2024Strength}) for fiber-reinforced structures with advanced multi-axis 3D printing techniques \cite{Liu2024NeuralSlicer} lies in representing fibers as vector fields. Converting these vector fields back into continuous toolpaths often introduces approximation errors, as vector fields are not always integrable (i.e., curl-free), limiting the ability to optimize them jointly. Our method proposed in this paper can effectively mitigate these issues while enabling co-optimization.

\subsubsection{Neural Based Representation}
Originally widely applied in computer vision for 3D reconstruction, \textit{implicit neural representations} (INRs) have recently gained traction in computational design and fabrication (e.g.,~\cite{Chandrasekhar2021TOUNN, woldseth2022use, Liu2024NeuralSlicer}). In these approaches, models or manufacturing processes are represented as INRs, with the coefficients of their neural networks optimized through loss function-driven self-learning by solvers such as Adam~\cite{kingma2014adam}.

In addition to INRs, many researchers have adopted data-driven strategy to tackle various optimization problems, including metamaterial design~\cite{Yue2023Neural}, iteration-free topology optimization~\cite{HU2024IFTONIR}, and toolpath planning ~\cite{Huang2024Learning}. However, a persistent challenge in data-driven methods is the acquisition of sufficient high-quality datasets for training. To address this limitation, \textit{physics-informed neural networks} (PINNs) offer an alternative by embedding prior knowledge and physics-based constraints, thereby reducing dependence on extensive datasets. PINNs have been successfully applied to problems such as fluid dynamics~\cite{RAISSI2019686}, topology optimization~\cite{JEONG2023116401}, and modeling composite manufacturing involving multi-physics~\cite{AKHARE2023Physics}. While PINNs demonstrate notable advantages in specific domains compared to traditional FEA, our experiments show that our current implementation based on FEA works effectively.
\section{Strength vs. Stiffness}\label{sec:Stiffness_Strength}

\subsection{Anisotropic Mechanical Properties}\label{subsec:AnisotropicProp}
\begin{figure}
\centering
\includegraphics[width=\linewidth]{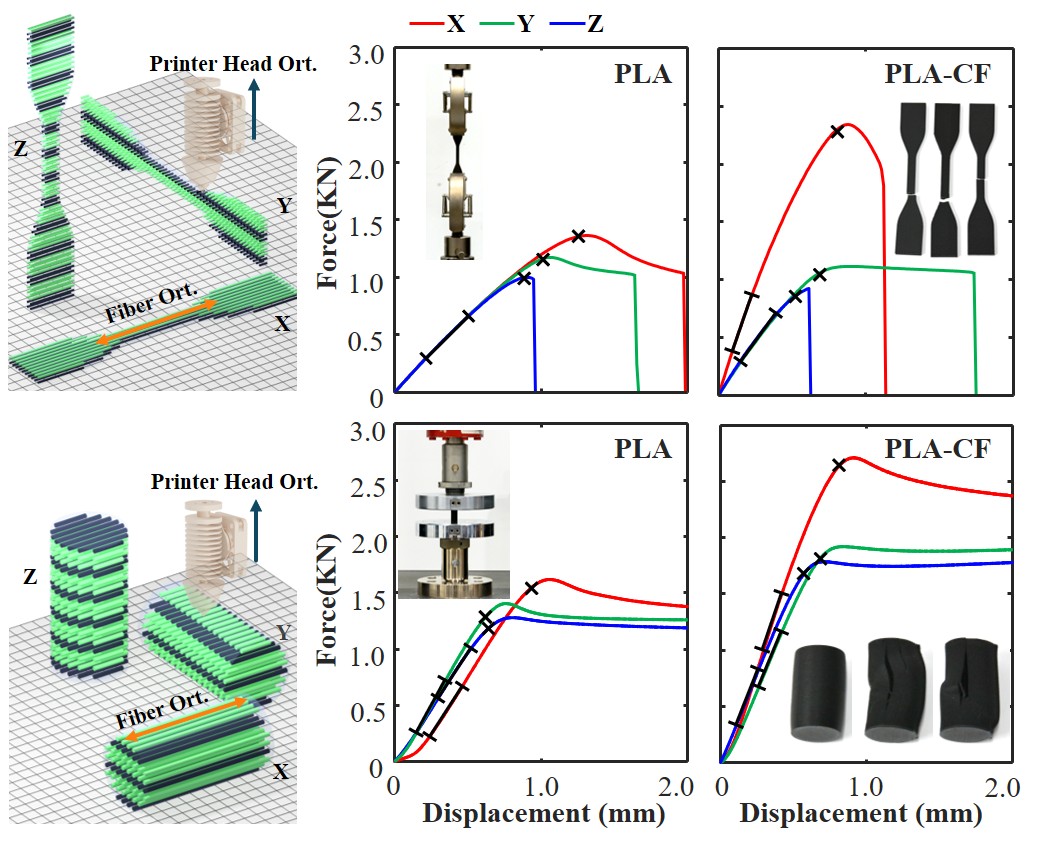}\\
\scriptsize
\begin{tabular}{r|c|c|c|c}
\hline
  & Young's  & \multicolumn{3}{c}{Strengths (Unit: MPa)}  \\
\cline{3-5}
Filament  & Modulus$^\dag$ (GPa) & Tensile & Compression & Shearing  \\
Type & $(E_x,E_y,E_z)$ & $(X_t,Y_t,Z_t)$  & $(X_c,Y_c,Z_c)$ & $(S_{xy},S_{yz},S_{zx})$  \\
\hline \hline
PLA & $(2.86,2.92,2.94)$ & $(42.4,31.8,27.9)$  & $(53.7,44.4,47.4)$ & $(24.2,15.4,20.2)$  \\
PLA-CF & $(7.68,3.24,3.16)$ & $(67.6,33.7,28.3)$ & $(85.2,61.3,63.3)$ & $(49.1,14.4,45.1)$  \\
\hline
\end{tabular}
\begin{flushleft}
\footnotesize
$^\dag$To simplify the numerical computation, an average of the tensile and compression moduli is used in our implementation.
\end{flushleft}
\caption{Strong anisotropic mechanical properties can be observed on specimens fabricated by filament-based 3D printing. In our tests, Young's modulus, tensile strength, compression strength, and shearing strength are measured by using the ASTM standards of E111, D638, D695, and D5379 respectively.}
\label{fig:anisotropicmech}
\end{figure}

Models fabricated through filament-based 3D printing exhibit pronounced anisotropy in mechanical properties~\cite{ahn2002anisotropic,tam20173d}. Fractographic analysis using \textit{scanning electron microscopy} (SEM)~\cite{Goldstein_introduction_sem} has identified weak adhesion between adjacent layers and incompletely filled regions between filaments as the primary causes of mechanical failure. This anisotropy in mechanical strength has been further validated by our experiments, including tensile and compression tests conducted on specimens printed in different orientations as shown in Fig.\ref{fig:anisotropicmech}. For a commonly used 3D printing material -- \textit{Polylactic Acid} (PLA) filament, its Young's modulus is almost isotropic (i.e., with <3\% variation in all directions). However, the yield strengths exhibit significant anisotropy due to the layered nature of filament-based 3D printing -- see the measurements of PLA as listed in Fig.\ref{fig:anisotropicmech}. All are measured on an INSTRON tensile machine 5960.

As a 3D printing filament that combines chopped carbon fiber with PLA, PLA-CF remains easy-to-print while offering superior strength and stiffness compared to regular PLA. The measurement shows the strong anisotropic Young's modulus of PLA-CF, which is $165\%$ higher than that of pure PLA. As given in Fig.\ref{fig:anisotropicmech}, its tensile strengths, compression strengths, and shearing strengths all present more significant anisotropy. These material properties, derived from experimental tests, will be used for the computation of FEA in our pipeline. Moreover, the measured anisotropic strengths will be employed to assess failure probabilities. 
 
Note that, although earlier work has shown that the level of anisotropy can be somewhat reduced by adjusting manufacturing parameters, such as extrusion rate and deposition force~\cite{Sun2023AMVoidReview}, strong anisotropy is still observed in PLA-CF even after these manufacturing adjustments. This aspect will not be further discussed here, as it is beyond the scope of this paper.

\subsection{Hoffman Criterion for Strength}\label{subsec:HoffmanCriterion}
As can be observed from Fig.\ref{fig:anisotropicmech}, the PLA-CF specimen exhibits an elastic response and fails in a brittle manner when subjected to both tensile and compression loads. Therefore, we disregard any plastic deformation of the filament under tension prior to yielding in this paper. Instead, we adopt the Hoffman criterion to evaluate the mechanical strength of composite structures. 

Given a stress $\boldsymbol{\sigma} \in \mathbb{R}^6$ defined in the material coordinate system by Voigt notation, the Hoffman yield criterion is formulated as a failure index in the quadratic form:
\begin{equation}\label{eq:HoffmanCriterion}
    \Gamma(\boldsymbol{\sigma}) = \boldsymbol{\sigma}^{T} \mathbf{Q} \boldsymbol{\sigma} + \mathbf{q}^T \boldsymbol{\sigma},
\end{equation}
where $\mathbf{Q} \in \mathbb{R}^{6 \times 6}$ is a symmetric and positive semi-definite matrix commonly determined by the strengths $(X_t,Y_t,Z_t)$, $(X_c,Y_c,Z_c)$ and $(S_{xy},S_{yz},S_{zx})$ (ref.~\cite{Schellekens1990}), and $\mathbf{q}\in \mathbb{R}^6$ is vector of linear coefficients as $(\frac{1}{X_t}-\frac{1}{X_c},\frac{1}{Y_t}-\frac{1}{Y_c},\frac{1}{Z_t}-\frac{1}{Z_c},0,0,0)$. According to the Hoffman criterion, the material failure occurs when $\Gamma(\boldsymbol{\sigma}) > 1.0$. We will explicitly evaluate it throughout the composite structure under optimization. 

\begin{figure}
    \centering
    \includegraphics[width=\linewidth]{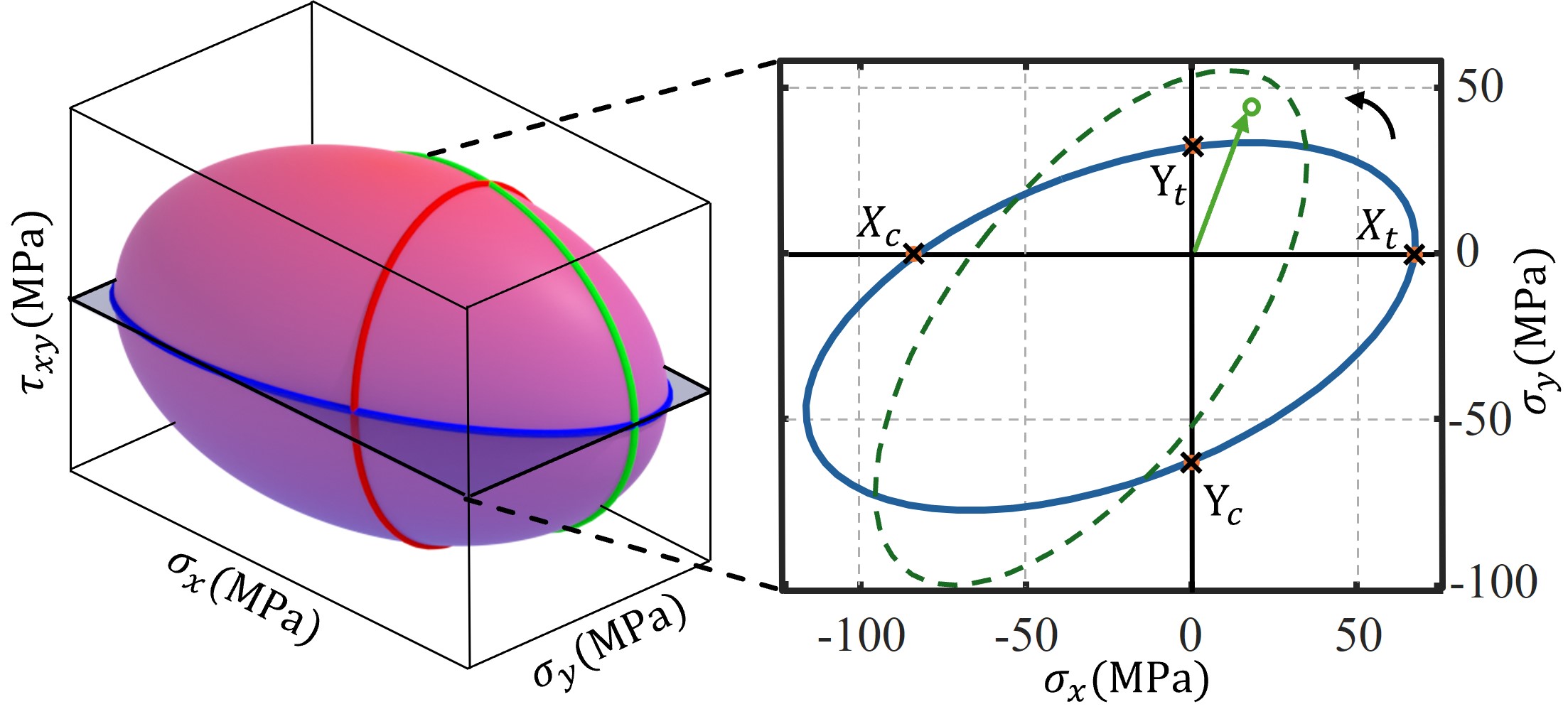}
    \put(-47,93){\small \color{black}$\boldsymbol{\sigma}_{d}$}
    \caption{The planar case with the tensile strengths $(X_t, Y_t)$ and the compression strengths $(X_c, Y_c)$ is employed to illustrate the Hoffman criterion ellipsoid. Re-orienting the anisotropic material effectively rotates the Hoffman criterion ellipsoid to enclose the yield point $\boldsymbol{\sigma}_{d}$, thereby preventing material failure under the given stress.
    }\label{fig:HoffmanCriterionEllipsoid}
\end{figure}

The quadratic form of the Hoffman criterion defines an ellipsoid in $\mathbb{R}^6$. For a yield point $\boldsymbol{\sigma}_d$ lies outside the ellipsoid, re-aligning the material orientations is conceptually equivalent to rotating the Hoffman criterion ellipsoid to encompass the point $\boldsymbol{\sigma}_d$. Figure~\ref{fig:HoffmanCriterionEllipsoid} illustrates this principle by a planar case with the stress components as $(\sigma_{x}, \sigma_{y}, \tau_{xy}) \in \mathbb{R}^6$. Nevertheless, it is important to note that adjusting the material orientations will also alter the stress distribution under a fixed load. 

Unlike the heuristic of aligning the fiber path with the principal stress directions conducted in previous approaches~\cite{Liu2024NeuralSlicer, Zhang2022s3, fang2020reinforced}, applying the Hoffman criterion to control the orientation of anisotropic materials offers more flexibility for filament-based 3D printing. For instance, in regions with low stresses, the fiber alignment can be more freely adjusted to ensure manufacturability, as long as the Hoffman criterion is satisfied. A comparison of the results obtained using these two strategies is presented in Fig.\ref{fig:ABracketNNSclier} of Sec.~\ref{subsubSec:SeqOptm_NeuralSlicer}, where our approach demonstrates the ability to generate stronger structures. 

\begin{figure}
\centering
\includegraphics[width=\linewidth]{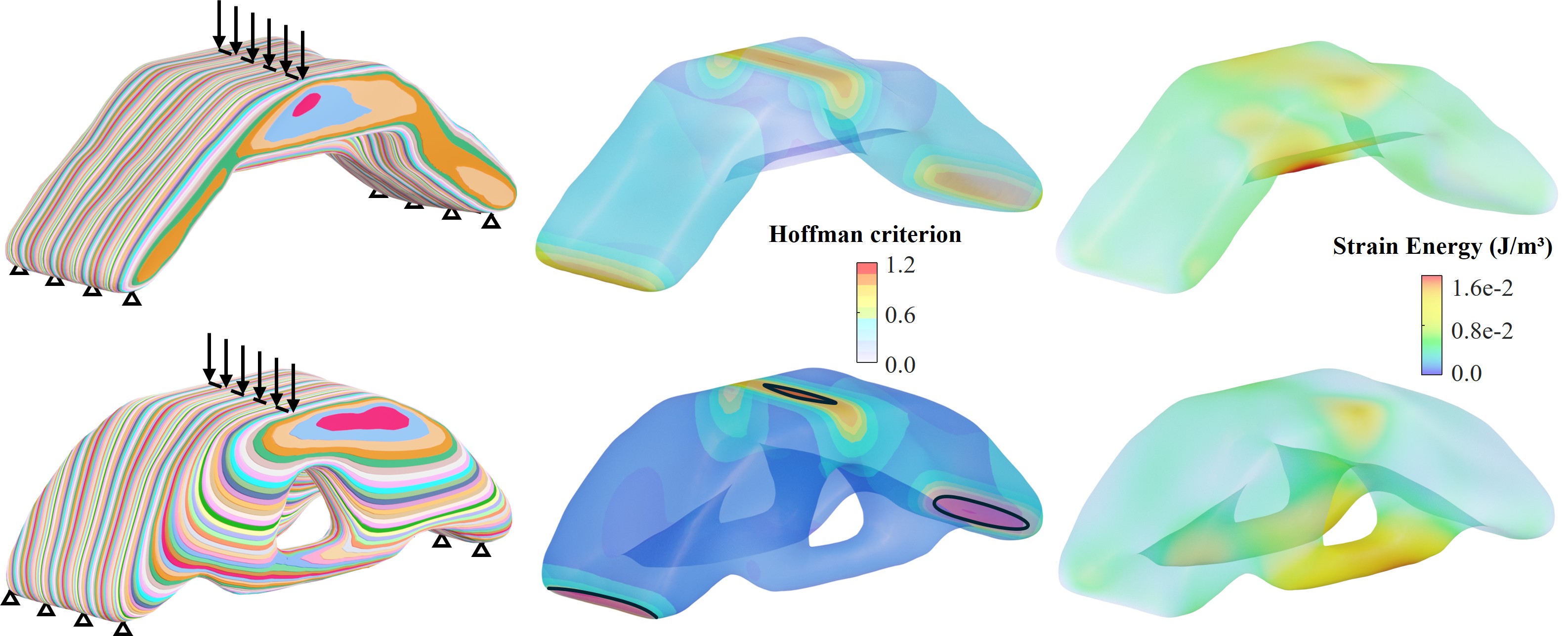}
\put(-210,55){\footnotesize \color{black}Strength-Based}
\put(-210,0){\footnotesize \color{black}Stiffness-Based}
\caption{An example of MBB beam with dimensions as $135 \times 45 \times 45 \, \text{mm} (w \times h \times d)$ having its structural topology optimized by the Hoffman criterion for strength (top row) and the compliance energy for stiffness (bottom row) -- both having manufacturability constraints incorporated and a target volume fraction of $25\%$. From left to right, the layers for multi-axis 3D printing, the failure index distribution $\Gamma(\cdot)$ and the strain energy distribution.
}\label{fig:resMBBStrengthStiffness}
\end{figure}

\subsection{Stiffness by Compliance}\label{subsec:Compliance}
Many studies in topology optimization assess a structure's stiffness by using compliance as a metric of its rigidity or flexibility \cite{Sigmund2001120}. For a structure under a load $\mathbf{F}$, the displacements $\mathbf{U}$ on all nodes of the structure can be determined by solving the equilibrium equation of FEA as $\mathbf{K}(\boldsymbol{\rho}) \mathbf{U} = \mathbf{F}$ with $\mathbf{K}(\boldsymbol{\rho})$ being a stiffness matrix depending on the density distribution $\boldsymbol{\rho}$ and also material orientations for anisotropic FEA. The compliance is defined as 
\begin{equation}\label{eq:Compliance}
    \Lambda(\boldsymbol{\rho}) =  \mathbf{U}^T \mathbf{K}(\boldsymbol{\rho}) \mathbf{U}.
\end{equation}
This compliance energy is often employed as the objective function to enhance structural performance by changing material distribution (i.e., the density field $\boldsymbol{\rho}$) inside the design domain. When anisotropic materials are employed, the stiffness matrix $\mathbf{K}(\boldsymbol{\rho})$ and the resultant displacements $\mathbf{U}$ depend on both the material distribution and the material orientations within the structure -- thus must be updated simultaneously.

Stiffness and strength optimizations target on different aspects of a structure when altering its topology. While stiffness-based optimization focuses on minimizing deformation under applied loads, it does not account for material failure explicitly. In contrast, strength-based optimization based on the Hoffman criterion is more appropriate for fiber-reinforced composites due to 
their application to provide strong mechanical strength, 
where failure is influenced by complex interactions among tensile, compressive, and shear stresses. This approach ensures the optimized structure can withstand realistic loading states without yielding. As illustrated by the comparison taken on the \textit{Messerschmitt-B\"{o}lkow-Blohm} (MBB) beam in Fig.\ref{fig:resMBBStrengthStiffness}, stiffness-optimized structures exhibit smaller strain energy (i.e., less deformation) but larger area of yielded regions (i.e., $\Gamma>1.0$), highlighting the limitations of stiffness-based methods for fiber-reinforced composites. 

\begin{figure*}
\includegraphics[width=\textwidth]{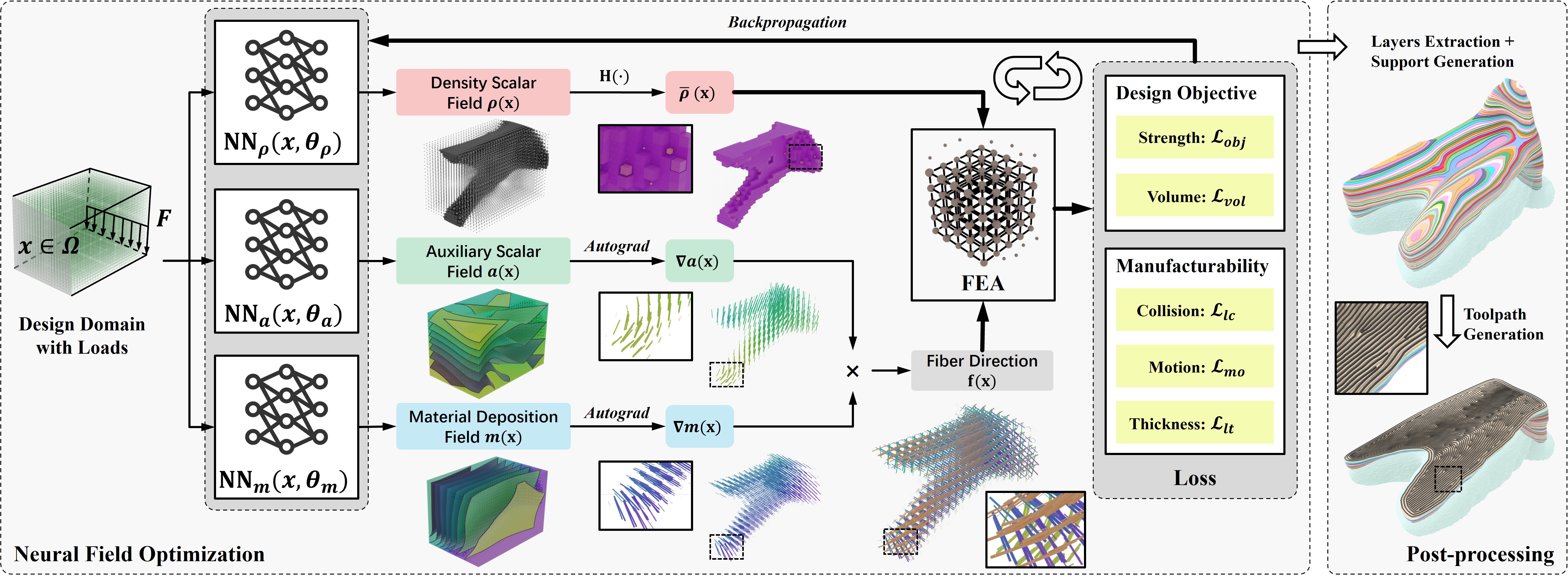}
\caption{Our computational pipeline for co-optimizing structural topology and manufacturing sequences employs a neural network-based representation. Structural solids, manufacturing layers (as the printing sequence), and fiber orientations (as the printing path direction) are represented by the density field $\rho(\mathbf{x},\theta_\rho)$, the material deposition field $m(\mathbf{x},\theta_m)$ and the auxiliary field $a(\mathbf{x},\theta_a)$ together. The network coefficients $\theta_\rho$, $\theta_m$ and $\theta_a$ serve as design variables and are optimized using backpropagation to minimize loss functions defined according to both design and manufacturing objectives, where the anisotropic FEA is employed to evaluate the mechanical performance for design objectives. 
}\label{fig:overview}
\end{figure*}

\section{Integrated Optimization}\label{sec:Overview}
Our computational pipeline for design and manufacturing integrated optimization of fiber-reinforced composites with strong anisotropy is based on the implicit neural representation, which will be first introduced in this section. After that, the overview algorithm of our co-optimization framework will be presented.

\subsection{Neural Field Based Representation}\label{subsec:NeuralFieldRep}
The basic idea of our approach is to construct a differentiable mapping as $\Phi: \forall \mathbf{x} \in\mathbb{R}^3 \mapsto (\rho,m,a) \in\mathbb{R}^3$ defined in the design domain $\Omega$, where the mapping is represented by three neural networks $\rho(\mathbf{x},\boldsymbol{\theta}_\rho)$, $m(\mathbf{x},\boldsymbol{\theta}_m)$ and $a(\mathbf{x},\boldsymbol{\theta}_a)$ as implicit neural fields. The fields are parameterized by the network coefficients $\boldsymbol{\theta}_\rho$, $\boldsymbol{\theta}_m$ and $\boldsymbol{\theta}_a$, and work for the following purpose (see also the illustration in Fig.~\ref{fig:overview}):
\begin{itemize}
\item \textbf{Topological Structure:} Inspired by the SIMP approach used in topology optimization \cite{Sigmund2001120}, a scalar density field $\rho(\mathbf{x},\boldsymbol{\theta}_\rho)$ as neural network is employed to represent the material distribution. At a given query point $\mathbf{x}\in\mathbb{R}^3$, $\rho(\mathbf{x},\boldsymbol{\theta}_\rho)=1$ indicates solid material and $\rho(\mathbf{x},\boldsymbol{\theta}_\rho)=0$ represents void, which are imposed by applying a sigmoid function $H(\cdot)$ as the continuous approximation of a Heaviside step function to filter the output of $\rho(\cdot)$.

\item \textbf{Manufacturing Sequence:} The second neural scalar field $m(\mathbf{x},\boldsymbol{\theta}_m)$ is to encode the order of material deposition, where the isosurfaces $\mathcal{G}_i$ as $m(\mathbf{x},\boldsymbol{\theta}_m)= c_i$ with $i=0,\ldots,n$ define the curved layers of multi-axis 3D printing. The preferred local printing direction and the layer thickness, are derived from the orientation and magnitude of the gradient $\nabla m(\mathbf{x})$. The curvature of each layer can also be evaluated on the implicit surface defined as $m(\mathbf{x})= c_i$.

\item \textbf{Fiber Orientations:} Since the distribution of fiber orientations $\mathbf{f}(\mathbf{x})$ as a vector field must align within the working surfaces, it cannot be defined independently. Instead, we introduce an auxiliary scalar field $a(\mathbf{x},\boldsymbol{\theta}_a)$ to define the fiber orientations together with $m(\mathbf{x},\boldsymbol{\theta}_m)$. Specifically, the fiber orientations can be computed by the product of gradients as $\mathbf{f}(\mathbf{x}) = \nabla a(\mathbf{x}) \times \nabla m(\mathbf{x})$, ensuring compatibility with the manufacturing sequence defined by $m(\cdot)$.
\end{itemize}  
When the networks are determined, we obtain all the information for evaluating the mechanical performance and the manufacturability of the composites under optimization.

\subsection{Anisotropic FEA}\label{subsec:AnisotropicFEA}
First of all, the geometry of a structure is defined as a set of points inside the design domain $\Omega$ with $\rho(\mathbf{x}) \simeq 1$, and the interpolated Young’s modulus of an element $e \in \Omega$ can be defined as
\begin{equation}
    E_e = E_{\min} +  H^p(\rho(\mathbf{x}_c^e)) (E_{\max} - E_{\min})
\end{equation}
in FEA formulation according to the density formulation of the SIMP method with $\mathbf{x}_c^e$ being the element $e$'s center and $p=3$ as the penalization factor \cite{Sigmund2001120}. $E_{\max}$ is the Young’s modulus of a solid element, and $E_{\min}$ is a small value to prevent the singularity of the stiffness matrix in FEA. The Young's modulus here has three components $(E_x,E_y,E_z)$ as analyzed in Sec.~\ref{subsec:AnisotropicProp}. 

The material coordinate frame is defined at any point $\mathbf{x}$ within the design domain as $(\mathbf{f}(\mathbf{x}), \nabla a(\mathbf{x}), \nabla m(\mathbf{x}))$. The transformation matrix $\mathbf{T}(\mathbf{x}) \in \mathbb{R}^{6\times6}$ that transforms a point from the material coordinate into the design coordinate -- both by Voigt notation -- can be derived to establish an element stiffness matrix as
\begin{equation}\label{eq:ElementStiffnessMat}
    \mathbf{K}_e = \mathbf{T}(\mathbf{x}_c^e) \mathbf{C}(E_e) (\mathbf{T}(\mathbf{x}_c^e))^{-1}
\end{equation}
with the orthotropic mechanical properties imposed in the constituent matrix $\mathbf{C}(E_e) \in \mathbb{R}^{6\times6}$. The system stiffness matrix of FEA, $\mathbf{K}_{\rho,m,a}$, is obtained by assembling all the element stiffness matrices. After that, the displacements $\mathbf{U}$ of the structure under optimization can be obtained by solving a linear equation system as 
\begin{equation}\label{eq:EquilibriumFEA}
    \mathbf{K}_{\rho,m,a} \mathbf{U} = \mathbf{F},
\end{equation}
where $\mathbf{F}$ is a given load. The compliance energy and the Hoffman criterion can then be evaluated by using the displacements $\mathbf{U}$ or the stresses as $\boldsymbol{\sigma} = \mathbf{C} \frac{\partial \mathbf{U}}{\partial \mathbf{x}}$. Details can be found in \cite{Schellekens1990}.
 
\subsection{Computational Pipeline}\label{subsec:CompPipeline}
The concurrent optimization of design and manufacturing for fiber-reinforced composites is formulated as solving the following problem with the help of FEA in a discrete form:
\begin{eqnarray}
    \label{eq:top}
            & \min & J_{des}(\boldsymbol{\theta}_\rho,\boldsymbol{\theta}_m,\boldsymbol{\theta}_a)   \\
         & s.t.& \mathbf{K}_{\rho,m,a} \mathbf{U} = \mathbf{F}, \nonumber \\
         & & \int_{\Omega} H(\rho(\mathbf{x})) \, d \mathbf{x} \leq V^*, \nonumber \\
         && \Pi_j(\boldsymbol{\theta}_\rho,\boldsymbol{\theta}_m,\boldsymbol{\theta}_a), \quad j=1,2,...,M. \nonumber
\end{eqnarray}
The objective function $J_{des}(\cdot)$ can be defined according to different design tasks -- the material failure condition based on the Hoffman criterion is employed for fiber-reinforced composites. The first constraint is the equilibrium equation (i.e., Eq.\eqref{eq:EquilibriumFEA}) for FEA computed on a discrete mesh to determine the displacements $\mathbf{U}$. The total volume of the structure is then evaluated in the second constraint and controlled by the maximally allowed volume $V^*$. The manufacturing requirements are defined as the constraints $\Pi_j(\cdot)$ in the last row, the details of which will be presented in Sec.~\ref{subsec:ManufacturingLoss}.  

We solve this optimization problem by a neural network-based learning pipeline as illustrated in Fig.~\ref{fig:overview}. The objective function and all the constraints are integrated into a total loss $\mathcal{L}_{total}$ to be minimized by backpropagation. Our algorithm consists of the following steps:
\begin{enumerate}
\item Discretize the design domain $\Omega$ into a mesh $\mathcal{M}$ for FEA; 

\item Initialize the neural networks $\rho(\mathbf{x},\boldsymbol{\theta}_\rho)$, $m(\mathbf{x},\boldsymbol{\theta}_m)$ and $a(\mathbf{x},\boldsymbol{\theta}_a)$;

\item For every element $e$, compute its element stiffness matrix $\mathbf{K}_e$ (Eq.~\eqref{eq:ElementStiffnessMat}) and fill it into the system stiffness matrix $\mathbf{K}_{\rho,m,a}$;

\item Solve Eq.~\eqref{eq:EquilibriumFEA} to determine the displacements $\mathbf{U}$;

\item Evaluate the loss functions according to design and manufacturing objectives;

\item Compute the gradients as $\frac{\partial \mathcal{L}_{total}}{\partial \boldsymbol{\theta}_{\rho}}$, $\frac{\partial \mathcal{L}_{total}}{\partial \boldsymbol{\theta}_{m}}$ and $\frac{\partial \mathcal{L}_{total}}{\partial \boldsymbol{\theta}_{a}}$;

\item Update the network coefficients $\boldsymbol{\theta}_{\rho}$, $\boldsymbol{\theta}_{m}$ and $\boldsymbol{\theta}_{a}$ by backpropagation; 

\item Go back to step (3) until the learning converges; 

\item Construct a volumetric mesh $\mathcal{M}^*$ for layer extraction;

\item Extract the isosurfaces of $m(\mathbf{x})$ on $\mathcal{M}^*$ and trim them by the implicit solid defined by $H(\rho(\mathbf{x}))$ to obtain the curved layers $\{\mathcal{G}_i\}$ (Sec.~\ref{subsec:PostProcessing});

\item Construct support structures by \cite{Zhang2023ICRA};

\item Generate toolpaths according to $\mathbf{f}(\mathbf{x})$ on each layer $\mathcal{G}_i$ by the method presented in \cite{zhang2024toolpath} (Sec.~\ref{subsec:PostProcessing}). 
\end{enumerate}
To achieve a balance between quality and efficiency, the volumetric mesh $\mathcal{M}^*$ for layer extraction has significantly higher resolution than the volumetric mesh $\mathcal{M}$ used for FEA. Both meshes are based on regular voxel grids. 

\section{Loss functions}\label{sec:LossFunc}
To optimize the structural topology, the design objectives related to strength and stiffness are first introduced and formulated as loss terms within our neural network-based computational pipeline. After that, manufacturing constraints are discussed and incorporated as additional loss terms. The total loss is then constructed by combining both the design and manufacturing losses.

\subsection{Design Objectives}\label{subsec:DesignObjectives}
\subsubsection{Strength} 
The design objective for strength-based optimization is defined using the concept of a safety factor $\gamma$, which represents the ratio between the capacity (i.e., the maximum allowable load before yielding) and the input load $\mathbf{F}$. The structural yield is evaluated by the Hoffman criterion as detailed in Sec.~\ref{subsec:HoffmanCriterion}. 

Given that $\boldsymbol{\sigma}_e$ is the stress on element $e$ obtained from FEA by solving Eq.~\eqref{eq:EquilibriumFEA} under the load $\mathbf{F}$, the stress on $e$ becomes 
$\gamma \boldsymbol{\sigma}_e$ when the maximum capacity (i.e., $\gamma \mathbf{F}$) is applied. This relationship holds true under the assumption of a linear elastic model in FEA. Specifically, solving $\mathbf{K}_{\rho,m,a} \mathbf{U}^* = \gamma \mathbf{F}$ and subsequently $\boldsymbol{\sigma}^* = \mathbf{C} \frac{\partial \mathbf{U}^*}{\partial \mathbf{x}}$ results in $\boldsymbol{\sigma}_e^* = \gamma \boldsymbol{\sigma}_e$. 

Inversely, the safety factor $\gamma_e$ for each element $e$ can be determined based on the Hoffman criterion as follows. By substituting $\boldsymbol{\sigma}^m = \gamma_e \boldsymbol{\sigma}_e$ into Eq.\eqref{eq:HoffmanCriterion}, the values of $\gamma_e$ are obtained as the positive roots of a quadratic equation $\Gamma(\gamma_e \boldsymbol{\sigma}_e) = A \gamma_e^2 + B \gamma_e  = 1$ 
with $A = \left(\boldsymbol{\sigma}_e\right)^T \mathbf{Q} \boldsymbol{\sigma}_e$ and $B = \mathbf{q}^T \boldsymbol{\sigma}_e$. Note that $A \geq 0$ is held here as $\mathbf{Q}$ is positive semi-definite. $A = 0$ only happens when $\boldsymbol{\sigma}_e \approx \mathbf{0}$, which gives $\gamma_e \rightarrow \infty$. In most other general cases, we have
\begin{equation}\label{eq:gammaE}
    \gamma_e = \frac{-B + \sqrt{B^2 + 4A}}{2A} > 0.
\end{equation}
The safety factor of the entire structure is determined by the minimum value of 
$\gamma_e$ across all elements. To enable a differentiable approximation of the minimum operator, we use the negative $p$-norm as a surrogate -- i.e., $\gamma = \min_e \{\gamma_e \} = ( \sum_{e} \gamma_e^{-\bar{p}} )^{-1/\bar{p}}$ as $\bar{p} \rightarrow \infty$ considering that $\gamma_e >0$. The design objective of strength-based optimization is to \textit{maximize} the value of $\gamma$.  Consequently, we define the loss function to be \textit{minimized} as: 
\begin{equation} \label{eq:LossStrength}
    \mathcal{L}_{obj} := - \left( \sum_{e} \gamma_e^{-\bar{p}} \right)^{-1/{\bar{p}}}.
\end{equation}
For numerical stability, we choose $\bar{p}=6$ in our implementation according to experiments.

\subsubsection{Volume} 
The volume of a structure directly relates to its weight, with the maximally allowed volume $V^*$ always being specified as an input of design optimization. To enforce this volume constraint as defined in Eq.\eqref{eq:top}, we introduce the following loss function:
\begin{equation} \label{eq:LossVolumn}
    \mathcal{L}_{vol} := ReLU \left(\frac{\sum_{e} H(\rho(\mathbf{x}_c^e)) V_e}{V^*} - 1 \right) 
\end{equation}
where $V_e$ is the volume of element $e$. The $ReLU(\cdot)$ function is used to let the loss only penalize the cases with volume exceeding $V^*$. 

\subsection{Manufacturability}\label{subsec:ManufacturingLoss}
The constraints ensuring the manufacturability of layers and fiber paths (i.e., $\{\Pi_j\}$ in Eq.\eqref{eq:top}) are defined as loss terms based on the field representation. The losses are evaluated at a set of points $\mathcal{P}$ uniformly sampling the whole design domain $\Omega$.  

\begin{figure}
\centering
\includegraphics[width=\linewidth]{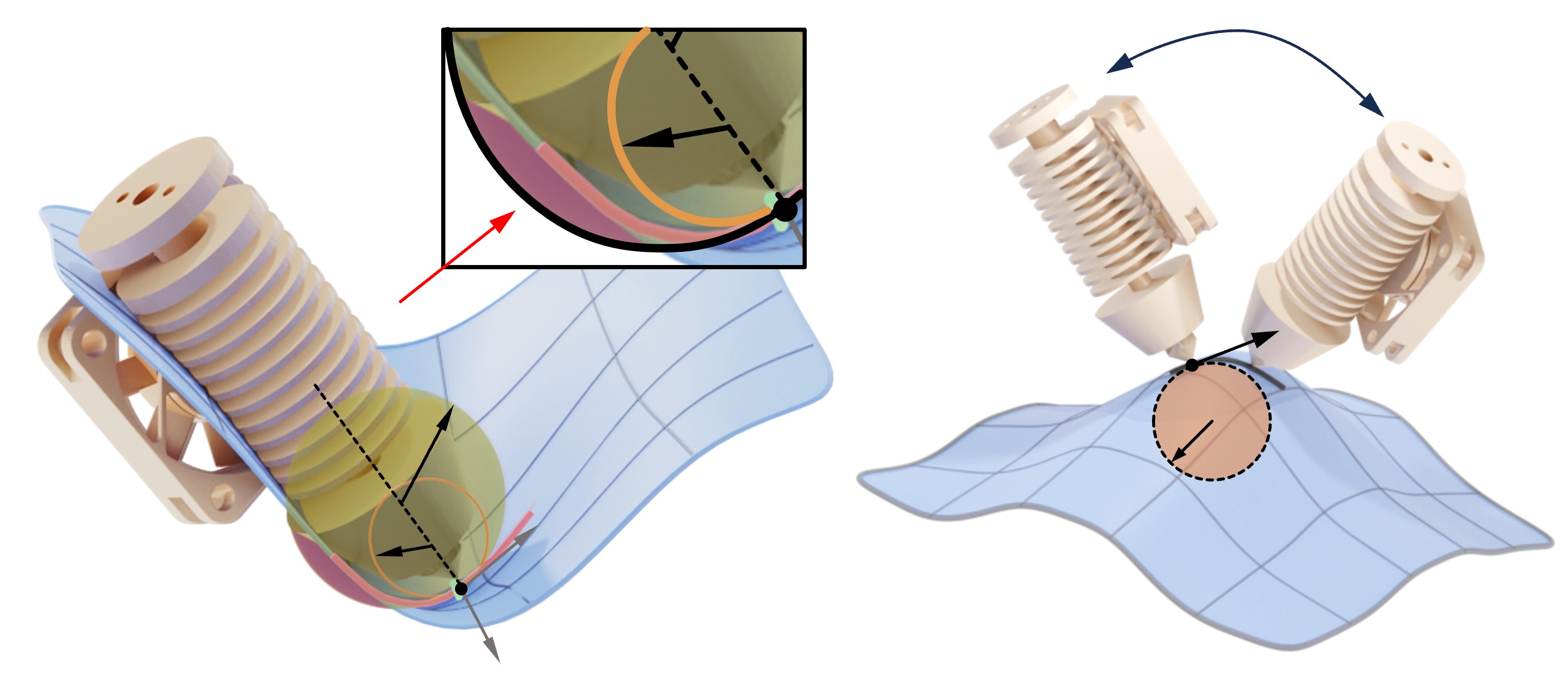}
\put(-240,5){\footnotesize \color{black}(a)}
\put(-110,5){\footnotesize \color{black}(b)}
\put(-180,49){\footnotesize \color{black}$K_{lc}^{-1}$}
\put(-203,25){\footnotesize \color{black}$K_{max}^{-1}$}
\put(-158,24){\footnotesize \color{black}$\boldsymbol{\kappa}_{max}$}
\put(-164,5){\footnotesize \color{black}$\boldsymbol{\kappa}_{min}$}
\put(-208,93){\footnotesize \color{black}Local}
\put(-208,85){\footnotesize \color{black}Collision}
\put(-45,56){\small \color{black}$\mathbf{f}(\mathbf{x})$}
\put(-72,27){\footnotesize \color{black}$K_f^{-1}$}
\caption{Illustrations of curvature based evaluations. (a) For a point $\mathbf{x}$ on the implicit surface defined by $m(\mathbf{x})=c_i$, whether there is local collision can be evaluated by the maximal curvature $K_{\max}$. Note that directions of the minimum curvature $\boldsymbol{\kappa}_{min}$ and the maximum curvature $\boldsymbol{\kappa}_{max}$ are specified by arrows. 
(b) The abrupt tool orientation change is prevented by controlling the curvature along the fiber-path direction $\mathbf{f}(\mathbf{x})$, denoted by $K_f$.
}\label{fig:CurvatureIllustration}
\end{figure}
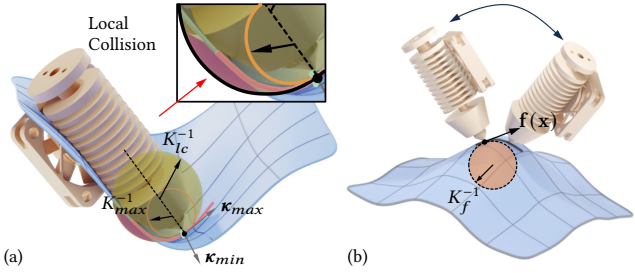

\subsubsection{Surface curvature for local collision}
Considering the axial symmetric shape of a printer head, the local collision constraint can be defined by limiting the allowed maximal curvature on a working surface $m(\mathbf{x}) = c_i$, which prevents the interference between the tool and the already printed parts of the model. A similar method was applied to avoid gougingin CNC machining  in~\cite{bartovn2021geometry} based on the parametric surface representation. In contrast, the mean and Gaussian curvatures are computed on implicit surfaces by the following formulas derived from~\cite{goldman2005curvature}:
\begin{equation}\label{eq:SurfCurvatureMean}
    K_M(\mathbf{x}) = \frac{ (\nabla m (\mathbf{x}))^{T} \, \mathbf{H}_m(\mathbf{x}) \, \nabla (m(\mathbf{x}) - \|\nabla m(\mathbf{x})\|^2 \text{tr}(\mathbf{H}_m(\mathbf{x}))}{2\|\nabla m (\mathbf{x})\|^3},
\end{equation}
\begin{equation}\label{eq:SurfCurvatureGaussian}
    K_G(\mathbf{x}) = \frac{(\nabla m (\mathbf{x}))^T \, \mathbf{H}^*_m(\mathbf{x}) \, \nabla m(\mathbf{x})}{\|\nabla m (\mathbf{x})\|^4},
\end{equation}
where $\nabla m(\mathbf{x})$ denotes the gradient of $m(\cdot)$ at a point $\mathbf{x} \in \mathbb{R}^3$, $\mathbf{H}_m(\mathbf{x})$ is the Hessian matrix of $m(\mathbf{x})$, and $\mathbf{H}^*_m(\mathbf{x})$ is the adjoint matrix of $\mathbf{H}_m(\mathbf{x})$. As $m(\mathbf{x})$ is defined by a neural network in our approach, both the first and the second derivatives can be analytically derived, based on which the maximum curvature of the implicit surface $m(x)=c_i$ can be obtained by 
\begin{equation}
    K_{\max}(\mathbf{x}) = K_M(\mathbf{x}) + \sqrt{(K_M(\mathbf{x}))^2 -K_G(\mathbf{x})}.
\end{equation}
Therefore, the local collision loss is defined as
\begin{equation}\label{eq:lossForCurvatureLC}
    \mathcal{L}_{lc} := \frac{1}{\Psi}\sum_{ \forall \mathbf{x}_p \in\mathcal{P} } H(\rho(\mathbf{x}_p)) ReLU(K_{\max}(\mathbf{x}_p) - K_{lc})
\end{equation}

to prevent the generation of non-manufacturable layers -- i.e., those with a maximum curvature greater than $K_{lc}$. We define the value of $K_{lc}$ based on the bounding sphere of the printer head (see Fig.\ref{fig:CurvatureIllustration} for an illustration). Note that this loss and all the losses in the rest of this sub-section only consider the sample points inside the structural solid with the help of $H(\rho(\cdot))$. The metric $\Psi$ measures the total accumulation of points inside the structural solid that can be computed as
\begin{equation}
    \Psi = \sum_{ \forall \mathbf{x}_p \in\mathcal{P} } {H(\rho(\mathbf{x}_p))}
\end{equation}
for normalizing the value of losses.

\subsubsection{Path curvature for motion control} Besides of $K_{\max}$ defined above to control the maximum curvature on all working surfaces, we also add a loss to avoid the abrupt tool orientation changes in motion by the directional curvature along the fiber-path direction $\mathbf{f}(\mathbf{x}) = \nabla a(\mathbf{x}) \times \nabla m(\mathbf{x})$. This can be evaluated from the normalized vector $\hat{\mathbf{f}} (\mathbf{x}) = \mathbf{f}(\mathbf{x}) / \| \mathbf{f}(\mathbf{x})\|$ as
\begin{equation}
    K_{f} := (\hat{\mathbf{f}}(\mathbf{x}))^T \, \mathbf{H}_f(\mathbf{x}) \hat{\mathbf{f}}(\mathbf{x}),
\end{equation}
where $\mathbf{H}_f(\mathbf{x})= \nabla \hat{\mathbf{f}}(\mathbf{x})$ is the Hessian matrix~\cite{goldman2005curvature}. As illustrated in Fig.\ref{fig:CurvatureIllustration}(b), this curvature directly relates to the derivative of tool orientation change. It can be controlled by the following loss
\begin{equation}
    \mathcal{L}_{mo} := \frac{1}{\Psi}\sum_{ \forall \mathbf{x}_p \in\mathcal{P} } H(\rho(\mathbf{x}_p)) ReLU(|K_f| - K_f^{\max})
\end{equation}

with $K_f^{\max}$ being a parameter that is determined by considering the machine's capability and the material property of filament to be printed. Note that this loss is only employed for 5-axis 3D printing as the tool orientation is unchanged during 3-axis and 2.5-axis printing.

\subsubsection{Setup orientation} 
For 5-axis 3D printing, the tool is able to be re-oriented along the surface normals (i.e., $\nabla m(\mathbf{x})$) during the process of material deposition. We employ a simple solution to define the setup orientation by the average of tool orientations.
\begin{wrapfigure}[8]{r}{0.32\linewidth}
\vspace{-10pt}
\centering
\hspace{-18pt}
\includegraphics[width=1.0\linewidth]{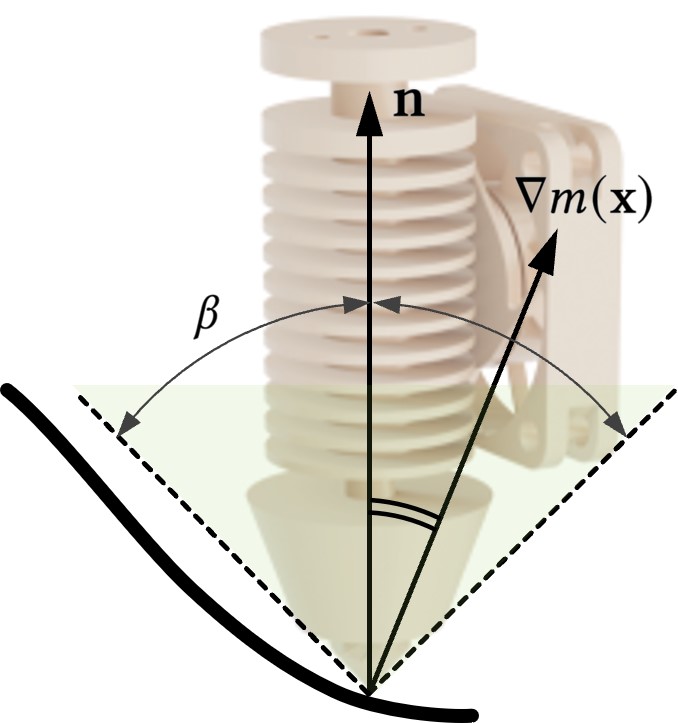}
\end{wrapfigure}

Differently, in 3-axis printing, we will need to determine the setup orientation of a model. This is defined as a printing orientation $\mathbf{n}$ in the model space to be optimized simultaneously, the value of which will also be updated via backpropagation of our computational pipeline. For 3-axis printing, to prevent the local collision, the maximal angle between the setup orientation $\mathbf{n}$ and the surface normal $\nabla m(\mathbf{x})$ needs to be constrained by the following loss
\begin{equation}
    \mathcal{L}_{ort} := \frac{1}{\Psi}\sum_{ \forall \mathbf{x}_p \in\mathcal{P} } H(\rho(\mathbf{x}_p)) ReLU\left( \frac{\nabla m (\mathbf{x}) \cdot \mathbf{n}}{\| \nabla m (\mathbf{x})\|\| \mathbf{n} \|} - \cos \beta \right),
\end{equation}

where $\beta$ is half of the apex angle at the printer head when assuming its shape as a cone with point contact to the working surface (see the insert above). 

For 2.5-axis printing, while optimizing the setup orientation $\mathbf{n}$ simultaneously, we also assign $\nabla m(\mathbf{x}) = \mathbf{n} / \| \mathbf{n} \|$ for every points within the design domain $\Omega$. Since only planar layers will be generated, the curvature losses $\mathcal{L}_{lc}$ and $\mathcal{L}_{mo}$ can be omitted, as they are always \textit{zero}. 

\subsubsection{Layer thickness}
Maintaining a consistent layer thickness is essential for achieving high-quality prints. The layer thickness loss ensures that the layers defined by $m(\cdot)$ adhere to the specified thickness, promoting printing uniformity and structural integrity. When requiring the layer thickness within the range of $[t_{\min}, t_{\max} ]$, we define the corresponding loss as 
\begin{eqnarray}
    \mathcal{L}_{lt} :=  \frac{1}{\Psi}\sum_{ \forall \mathbf{x}_p \in\mathcal{P} } H(\rho(\mathbf{x}_p)) (ReLU( t_{min} - \| \nabla m (\mathbf{x}) \|) \nonumber \\
    + ReLU(\| \nabla m (\mathbf{x}) \| - t_{max})).
\end{eqnarray}

As a byproduct, the layer thickness loss can also effectively avoid the vanished gradient on $m(\mathbf{x})$. This improves the numerical stability of other losses with $\| \nabla m(\mathbf{x}) \|^{-1}$ term.

\subsection{Total Loss}
The total loss function for the co-optimization of composites to be printed on a 5-axis machine is defined by the weighted sum of the relevant losses as
\begin{equation}\label{eq:TotalLoss}
    \mathcal{L}^{\text{5x}}_{total} := \omega_{obj}\mathcal{L}_{obj} +  \omega_{vol}\mathcal{L}_{vol}+ \omega_{lc}\mathcal{L}_{lc} + \omega_{mo}\mathcal{L}_{mo} + \omega_{lt}\mathcal{L}_{lt}.
\end{equation}

Differently, composites to be fabricated by a 3-axis printer is optimized using a loss as
\begin{equation}\label{eq:TotalLoss_3x}
    \mathcal{L}^{\text{3x}}_{total} := \omega_{obj}\mathcal{L}_{obj} +  \omega_{vol}\mathcal{L}_{vol}+ \omega_{lc}\mathcal{L}_{lc} + \omega_{ort}\mathcal{L}_{ort} + \omega_{lt}\mathcal{L}_{lt},
\end{equation}
where the loss $\mathcal{L}_{mo}$ for 5-axis motion is replaced by the setup orientation loss $\mathcal{L}_{ort}$ for 3-axis printing. The total loss function for planar printing is much simpler as the field $m(\cdot)$ is degenerated into a field with constant gradient as the orientation of all layers specified by $\mathbf{n}$. It only has the design objective and the volume requirement left as
\begin{equation}\label{eq:TotalLoss_planar}
    \mathcal{L}^{\text{planar}}_{total} := \omega_{obj}\mathcal{L}_{obj} +  \omega_{vol}\mathcal{L}_{vol}.
\end{equation}
That means there is no additional manufacturing constraint after imposing the planar fiber orientations as $\mathbf{f}(\mathbf{x})= \nabla a(\mathbf{x}) \times \mathbf{n}$.

The weights of loss terms are chosen dynamically to achieve a balance between the design objective loss and the other losses as constraints. The detailed scheme of weighting can be found in Sec.~\ref{subsec:WeightingScheme}.

\section{Implementation Details}\label{sec:DetailImplementation}

\subsection{Differentiable Pipeline}\label{subsec:Diff}
Based on our formulation presented in Sec.~\ref{sec:LossFunc}, all loss functions are differentiable so that they can be effectively optimized by gradient-based solvers. The analysis is given below. 

We first consider the design objective $\mathcal{L}_{obj}$. Given $\theta$ as any scalar variable among all network coefficients $\boldsymbol{\theta}_{\rho}$, $\boldsymbol{\theta}_m$ and $\boldsymbol{\theta}_a$, we have
\begin{equation}
\frac{\partial \mathcal{L}_{obj}}{\partial \theta} 
    = \frac{1}{\bar{p}}\left( \sum_{e} \gamma_e^{-\bar{p}} \right)^{-\frac{1+\bar{p}}{\bar{p}}} 
    \sum_{e} -\bar{p} \gamma_e^{-(\bar{p}+1)} 
    \frac{\partial \gamma_e}{\partial \theta} 
\end{equation}
with $\partial \gamma_e / \partial \theta$ being
\begin{equation}
    \frac{\partial \gamma_e}{\partial \theta} = 
    \frac{\partial \gamma_e}{\partial A} \left( \mathbf{Q}^T \boldsymbol{\sigma}_e \cdot \frac{\partial \boldsymbol{\sigma}_e}{\partial \theta} \right) +
    \frac{\partial \gamma_e}{\partial B} \left( \mathbf{q}^T \cdot \frac{\partial \boldsymbol{\sigma}_e}{\partial \theta}  \right)
\end{equation}
obtained from Eq.\eqref{eq:gammaE}. Since $\boldsymbol{\sigma}_e$ is defined by the element displacement vector $\mathbf{u}_e$ as 
$\boldsymbol{\sigma}_e = \mathbf{C} \mathbf{T}_e^{T} \frac{\partial \mathbf{u}_e}{\partial \mathbf{x}}  \mathbf{T}_e$, we have
\begin{equation}
   \frac{\partial \boldsymbol{\sigma}_e}{\partial \theta}=
   \mathbf{C}\frac{\partial \mathbf{T}_e^{T}}{\partial \theta}  \frac{\partial \mathbf{u}_e}{\partial \mathbf{x}}   \mathbf{T} + 
   \mathbf{C}\mathbf{T}^{T}_e \frac{\partial^2 \mathbf{u}_e}{\partial \mathbf{x}\partial \theta} {\partial \mathbf{x}} \mathbf{T}_e + 
   \mathbf{C}\mathbf{T}_{e}^T \frac{\partial \mathbf{u}_e}{\partial \mathbf{x}}  \frac{\partial \mathbf{T}_e} {\partial \theta}.
\end{equation}
Here $\mathbf{T}_{e}$ is a differentiable rotation matrix, which only relies on the neural fields $m(\cdot)$ and $a(\cdot)$.

The system displacement $\mathbf{U}$ as an assembly of $\{ \mathbf{u}_e \}$ is determined by FEA according to $ \mathbf{K} \mathbf{U} = \mathbf{F}$. Therefore, we have
\begin{equation}\label{eq:PartialU_PartialTheta}
    \frac{\partial \mathbf{U}}{\partial \theta} =  \frac{\partial (\mathbf{K}^{-1}\mathbf{F})}{\partial \mathbf{\theta}} = - \mathbf{K}^{-1} \mathbf{F} \otimes \mathbf{K}^{-1} \frac{\partial \mathbf{K}}{\partial \theta}
\end{equation}
by using the Kronecker product $\otimes$. As the system stiffness matrix $\mathbf{K}$ is commonly determined by $\rho(\cdot)$, $m(\cdot)$ and $a(\cdot)$ at the center points $\{ \mathbf{x}^e_c\}$ of all elements, it is differentiable in terms of the network coefficient $\theta$.  
When evaluating these gradients, we do not need to explicitly compute the inverse of a sparse matrix $\mathbf{K}$. Alternatively, the linear systems $\mathbf{K} \mathbf{U} = \mathbf{F}$ and $\mathbf{K} \mathbf{Y} = \frac{\partial \mathbf{K}}{\partial \theta}$ are solved to obtain $\mathbf{U}$ and $\mathbf{Y}$ as $\mathbf{K}^{-1} \mathbf{F}$ and $\mathbf{K}^{-1} \frac{\partial \mathbf{K}}{\partial \theta}$ for Eq.\eqref{eq:PartialU_PartialTheta}, the computation of which is more efficient in practice.

The volume loss $\mathcal{L}_{vol}$ is explicitly defined by the neural field function $\rho(\cdot)$. Consequently, its differentiation $\partial \mathcal{L}_{vol} / \partial \theta_\rho$ can be directly obtained from the network, while $\partial \mathcal{L}_{vol} / \partial \theta_m = \partial \mathcal{L}_{vol} / \partial \theta_a = 0$. Additionally, all manufacturability losses, including $\mathcal{L}_{lc}$, $\mathcal{L}_{mo}$ and $\mathcal{L}_{lt}$ are explicitly dependent on $\rho(\cdot)$, $m(\cdot)$ and $a(\cdot)$ and thus on their respective network coefficients $\boldsymbol{\theta}_{\rho}$, $\boldsymbol{\theta}_m$ and $\boldsymbol{\theta}_a$. Similarly, the setup orientation loss $\mathcal{L}_{ort}$ for 3-axis printing is explicitly defined by $\boldsymbol{\theta}_{\rho}$, $\boldsymbol{\theta}_m$,  $\boldsymbol{\theta}_a$ and $\mathbf{n}$. In all cases, the analytical form of differentiation with respect to the network coefficients exists. 

\subsection{Network Architecture and Learning}
The architecture of a \textit{multilayer perceptron} (MLP) is utilized to model each of three scalar fields $\rho(\cdot)$, $m(\cdot)$, and $a(\cdot)$. Our implementation consists of 5 hidden layers each containing 256 neurons, which is chosen experimentally to effectively capture the complexity of these fields. 

To ensure third-order continuity, improve differentiability, and facilitate better convergence, the \textit{Sigmoid Linear Unit} (SiLU) activation function is employed across all layers. The networks take the position vector $\mathbf{x} \in \mathbb{R}^3$ as input and output the corresponding values of the scalar fields $\rho(\mathbf{x})$, $m(\mathbf{x})$ and $a(\mathbf{x})$ respectively.

We implement the proposed co-optimization pipeline using the PyTorch framework~\cite{Adam2019PyTorch}, leveraging its NN-based machine learning capabilities. This allows us to minimize the loss function effectively by combining automatic gradient computation with a NN solver (i.e., Adam optimizer~\cite{kingma2014adam}), which updates the network coefficients through backpropagation. 
During the step of network initialization, the first four MLP layers are randomly initialized. Specific initial values are assigned for the last layer to facilitate meaningful starting conditions: 1) uniform values of $0.5$ for the density field $\rho(\cdot)$, 2) uniform gradients along the $z$-axis for the material deposition field $m(\cdot)$, and 3) uniform gradients along the $y$-axis for the auxiliary field $a(\cdot)$.
The optimization process starts with an initial learning rate of $1.0e-3$, which is adaptively adjusted during training using the `ReduceLROnPlateau' scheduler. This scheduler monitors the optimization progress and dynamically reduces the learning rate when a plateau is detected, promoting efficient convergence. The minimum learning rate threshold is set to $1.0e-6$ to better control the convergence.

\subsection{Weighting Scheme}\label{subsec:WeightingScheme}
The optimization problem with integrated design and manufacturing requirements is in general complicated. Therefore, multiple constraints are defined together with the design objective as multiple loss terms in the total loss to be minimized. We implement a hybrid weighting scheme, combining the penalty method from the TOuNN framework \cite{Chandrasekhar2021TOUNN} and the DC3 framework \cite{donti2021dc3}, to ensure those constraints are effectively preserved.
\begin{itemize}
\item First, we pre-compute the unconstrained problem by only using the $\mathcal{L}_{obj}$ term and set $\omega_{obj}=10/\mathcal{L}_{obj}$. After that, the weight $\omega_{obj}$ for the design objective is dynamically adjusted to balance the value of $\omega_{obj} \mathcal{L}_{obj}$ with other terms in the total loss, keeping its value in the range of $[0, 10]$. This can effectively normalize the design objective term during optimization.

\item Second, the weights of all other losses are initialized by zero value and increased by $0.05$ at each step, ensuring a smooth transition toward enforcing constraints while maintaining convergence stability.  

\item Last but not the least, a gradient-based correction step is applied to all constraints when $\mathcal{L}_{obj}$ is less than 1.0e-5, ensuring that solutions are progressively aligned closer to the feasible region. This correction step is realized by assigning $\omega_{obj}=0.0$ when applying backpropagation to update the network coefficients.
\end{itemize}
This hybrid approach ensures that the optimization process satisfies those manufacturing constraints while adapting dynamically to the changes in the design objective function.

\begin{table*}[t]
\caption{Statistics for our co-optimization pipeline.}\vspace{-5pt}
\centering\label{tabCompStatistic}
\footnotesize
\begin{tabular}{l l c r || r r r r | r || c | c | c  || c | r }
\hline 
& & & & \multicolumn{4}{c|}{Computing Time (sec.)}  & Total & \multicolumn{3}{c||}{Fabrication Info.} & \multicolumn{2}{c}{Memory (GB)} \\
\cline{5-8}
\cline{10-12} 
\cline{13-14} 
Model & Method & Fig.  &FEA Grid Res.  & FEA$^\dag$ & Optm.$^\dag$  & Slicing & Path Gen. & Time (sec.) &\#Layer & Weight (g) & Time (h) & RAM & VRAM\\
\hline \hline
GE-Bracket & CoOptm  & \ref{fig:teaser} & $92 \times 31 \times 51$ & $18.89$ & $2.42$ & $114.75$ & $205.21$ & $29,893$ & $150$ & $260.5$ & $33.6$& $43.59$& $15.47$\\
GE-Bracket & SeqOptm  & \ref{fig:teaser}  & $92 \times 31 \times 51$ & $18.89$ & $2.42$ & $121.33$ & $234.31$ & $71,409$ & $172$ & $261.3$ & $35.2$ & $43.59$ & $15.47$\\
GE-Bracket & CoOptm (3-Axis)  & \ref{fig:teaser}  & $92 \times 31 \times 51$ & $18.89$ & $1.85$ & $119.36$ & $219.16$ & $22,762$ & $158$ & $259.1$ & $25.8$ & $42.36$ & $13.58$\\
GE-Bracket & CoOptm (2.5-Axis)  & \ref{fig:teaser}  & $92 \times 31 \times 51$ & $18.89$ & $1.15$ & $37.5$ & $103.23$ & $18,195$ & $129$ & $261.2$ & $20.9$ & $29.35$ & $9.72$\\
MBB Beam & CoOptm & \ref{fig:resMBBStrengthStiffness}  & $45 \times 15 \times 15$ & $5.35$ & $0.54$ & $56.64$ & $97.2$ & $2,275$ & $76$ & $/$ & $/$ & $10.60$ & $3.561$\\
Cantilever  & CoOptm & \ref{fig:overview}  & $45 \times 30 \times 30$ & $7.42$ & $0.85$ & $65.41$ & $86.5$ & $3,762$ & $83$ & $/$ & $/$ & $15.83$ & $5.713$\\
L-Bracket & CoOptm  & \ref{fig:Lbracket}  & $50 \times 50 \times 50$ & $16.65$ & $2.21$ & $105.41$ & $168.72$ & $15,156$ & $125$ & $/$ & $/$ & $40.67$ & $15.68$\\
L-Bracket & SeqOptm  & \ref{fig:Lbracket}  & $50 \times 50 \times 50$ & $16.65$ & $2.21$ & $110.95$ & $194.41$ & $21,091$ & $142$ & $/$ & $/$ & $40.67$ & $15.68$\\
A-Bracket & CoOptm & \ref{fig:ABracketNNSclier}, \ref{fig:ABracketL2}  & $43 \times 39 \times 49$ & $10.47$ & $1.17$ & $100.25$ & $173.31$ & $10,406$ & $126$ & $/$ & $/$ & $40.67$ & $14.20$\\
\hline
\end{tabular}
\begin{flushleft}
$^\dag$~The computing time of FEA and the rest of optimization is reported as the average value of each step.
\end{flushleft}
\end{table*}

\subsection{Post-processing: Slicing and Toolpath Generation}\label{subsec:PostProcessing}
The curved layers for multi-axis 3D printing are generated by extracting the isosurface surfaces of the material deposition field $m(\mathbf{x})$, using the \textit{Marching Cube} (MC) algorithm~\cite{Lorensen1987Marching} on a voxel grid mesh $\mathcal{M}^*$ at the resolution of $256^3$ defined in the design domain $\Omega$. The isovalues are chosen according to the required layer thickness of printing. After obtaining the triangular mesh surfaces from MC, they are trimmed by the implicit solid $H(\rho(\mathbf{x})) \leq 0.5$ and then remeshed to form the curved layers $\{ \mathcal{G}_i \}$. Based on these layers, the supporting structures are generated by the method presented in \cite{Liu2024NeuralSlicer,Zhang2023ICRA}. 

The toolpaths for printing fiber-reinforced composites are generated on each curved layer $\mathcal{G}_i$. Ideally, the toolpaths are designed to align with the fiber field 
$\mathbf{f}(\mathbf{x}) = \nabla a(\mathbf{x}) \times \nabla m(\mathbf{x})$, which is analytically defined at any point on the surface $\mathcal{G}_i$. Unlike existing approaches (e.g.,~\cite{fang2020reinforced, Luo2023Spatially}) that directly define the toolpaths as iso-curves of a scalar field derived via discrete Hodge decomposition of the vector field, we employ a 2-RoSy representation of the direction field with respect to $\pm \mathbf{f}(\mathbf{x})$ and a periodic scalar field generated through periodic parameterization, the strategy which is also used in \cite{Knoppel2015Stripe, zhang2024toolpath}. This approach generates fiber toolpaths as partial iso-curves of the periodic scalar field with nearly uniform hatching distances. This method ensures that the toolpaths can closely follow the directions of $\pm \mathbf{f}(\mathbf{x})$ while avoiding distortions caused by singularities in the vector field. 
\section{Results}

\subsection{Computational Experiments}
Our computational pipeline is implemented in Python, while the post-processing code of toolpath and support generation are based on the implementations in C++. The source code of our pipeline will be released upon the acceptance of this paper. All computational experiments are performed on a desktop PC with an Intel Core i5-12600K CPU (10 cores @ 3.6GHz) with 96GB RAM and an NVIDIA RTX4080 GPU with 16GB VRAM, running Ubuntu 20.04 LTS. 

\subsubsection{Statistics for computational efficiency} 
The computational statistics for all examples tested in our experiments are summarized in Table \ref{tabCompStatistic}. It can be observed that the optimization process in our pipeline is completed within 38 minutes to 8 hours, except for the sequential optimization applied to the GE-Bracket, which takes approximately 20 hours due to the two phases of iterations. The most time-consuming step in our computational pipeline is the FEA, even though we have employed relatively low-resolution grid meshes to reduce computational cost. During optimization, the loss functions for all examples are evaluated on a set of $N$ points uniformly sampled within the design domain, where $N$ is twice the number of grid elements used in the FEA. The table also lists the number of resultant layers, as well as the usage of main memory (RAM) and graphics memory (VRAM).

\subsubsection{Verification by FEA}\label{subsubSec:FEA}
The mechanical strength of resultant fiber-reinforced composites can be verified by the yield analysis performed using commercial FEA software, specifically Abaqus in our experiments. First of all, the surface model of a resultant structure is extracted from the isosurface $H(\rho(\mathbf{x})) = 0.5$ by the MC algorithm, which is thereafter remeshed and employed to generate a tetrahedral mesh by TetGen~\cite{hang2015tetgen}. For each element $e$ of the tetrahedral mesh, the element's material coordinate frame can be assigned according to $(\mathbf{f}(\mathbf{x}^e_c), \nabla a(\mathbf{x}^e_c), \nabla m(\mathbf{x}^e_c))$ at the center of the element, $\mathbf{x}^e_c$. The volume meshes together with material coordinate frames are then imported into Abaqus to conduct anisotropic FEA according to the specified loads and the anisotropic material properties of PLA-CF as discussed in Sec.\ref{subsec:AnisotropicProp}. After that, the resultant stress tensors in all elements are exported to compute the values of Hoffman yield criterion $\Gamma(\cdot)$ on every elements. The examples of failure index distributions are shown in Figs.\ref{fig:teaser} and \ref{fig:resMBBStrengthStiffness}. The strain energy distribution as shown in the right column of Fig.\ref{fig:resMBBStrengthStiffness} can also be generated by the strains obtained from the anisotropic FEA.

\subsubsection{Concurrent vs. sequential optimization} 
We now study the advantages of our concurrent optimization approach. The results are compared to those obtained from a sequential approach that optimizes the structural topology and the manufacturable layers separately. Specifically, for generating the Phase I result of sequential optimization, we assign $\omega_{lc}$, $\omega_{mo}$ and $\omega_{lt}$ in Eq.~\eqref{eq:TotalLoss} as \textit{zero} to compute the structural topology by $\rho(\cdot)$ while obtaining the fiber-orientations by $a(\cdot)$ and $m(\cdot)$ -- i.e., using design objectives only. After that, the Phase II optimization is conducted by locking the network coefficients $\boldsymbol{\theta}_{\rho}$ but using the full loss function defined in Eq.~\eqref{eq:TotalLoss} with nonzero weights. Therefore, the structural topology is unchanged in Phase II.

\begin{figure}
\centering
\includegraphics[width=\linewidth]{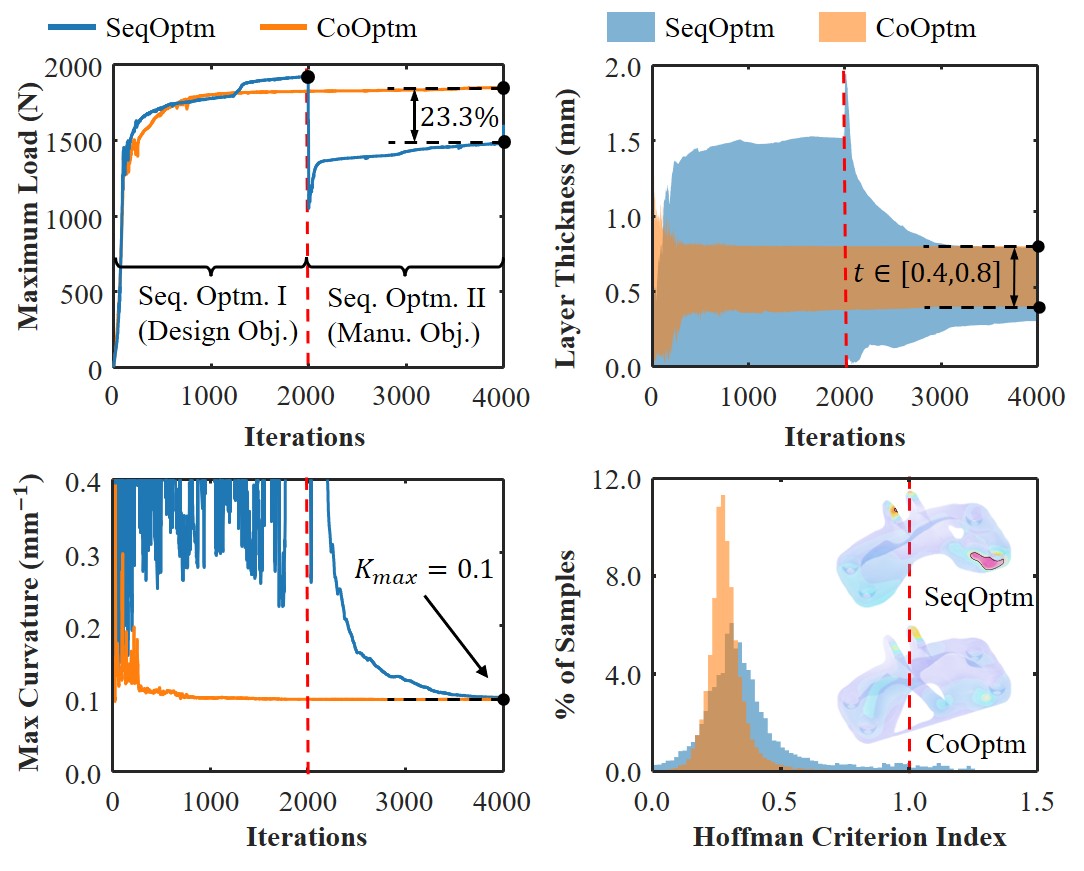}\vspace{-8pt}
\put(-232,103){\footnotesize \color{black}(a)}
\put(-110,103){\footnotesize \color{black}(b)}
\put(-232,9){\footnotesize \color{black}(c)}
\put(-110,9){\footnotesize \color{black}(d)}
\caption{Comparison of our co-optimization approach vs. the sequential optimization: (a) evolution of maximum load during the optimization process; (b) changes in layer thickness with the required range as $[0.4,0.8]$; (c) variation of maximal curvatures on layers; (d) the histograms of the Hoffman criterion values in all elements.}
\label{fig:GEBracketL2_1StageVS2StageTouopt}
\end{figure}
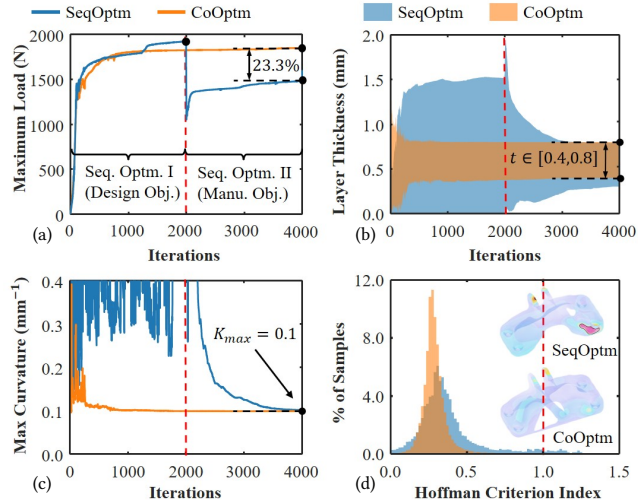

The first example tested in our study is the GE-Bracket as shown in Fig.\ref{fig:teaser}, which includes four bolt holes for securing the structure and two through-holes for horizontal loads. The design domain has the dimension as $200 \times 110 \times 70 \, \text{mm} \; (w \times h \times d)$, and the FEA in optimization was computed by using a voxel grid mesh with voxel size as $2\,\mathrm{mm}$. 
As already visualized in Fig.~\ref{fig:teaser}, the result of sequential optimization has large yielded areas when applying the load $F=1.855~\text{kN}$ as the maximum yield-free force obtained from the co-optimization. The convergence curves of maximum loads for sequential vs. concurrent optimizations are given in Fig.~\ref{fig:GEBracketL2_1StageVS2StageTouopt}(a). The maximum yield-free load $F_{yd}$ achieved by our co-optimization is $3.6\%$ 
lower than the Phase I result of sequential optimization, which is not manufacturable. However, ours is $23.3\%$ higher after incorporating the manufacturability in Phase II of the sequential optimization while keeping the structural topology unchanged. Comparisons of layer thickness and maximum curvature are illustrated in Fig.~\ref{fig:GEBracketL2_1StageVS2StageTouopt}(b) and \ref{fig:GEBracketL2_1StageVS2StageTouopt}(c) respectively. 
It shows that the layer thickness requirements remain unmet in Phase II of the sequential optimization. Additionally, histograms of Hoffman criterion indices for all elements are shown in Fig.~\ref{fig:GEBracketL2_1StageVS2StageTouopt}(d), demonstrating that the index values for the co-optimization results are all below $1.0$ when applying the load $F=1.855~\text{kN}$. 

\begin{figure}
\centering
\includegraphics[width=\linewidth]{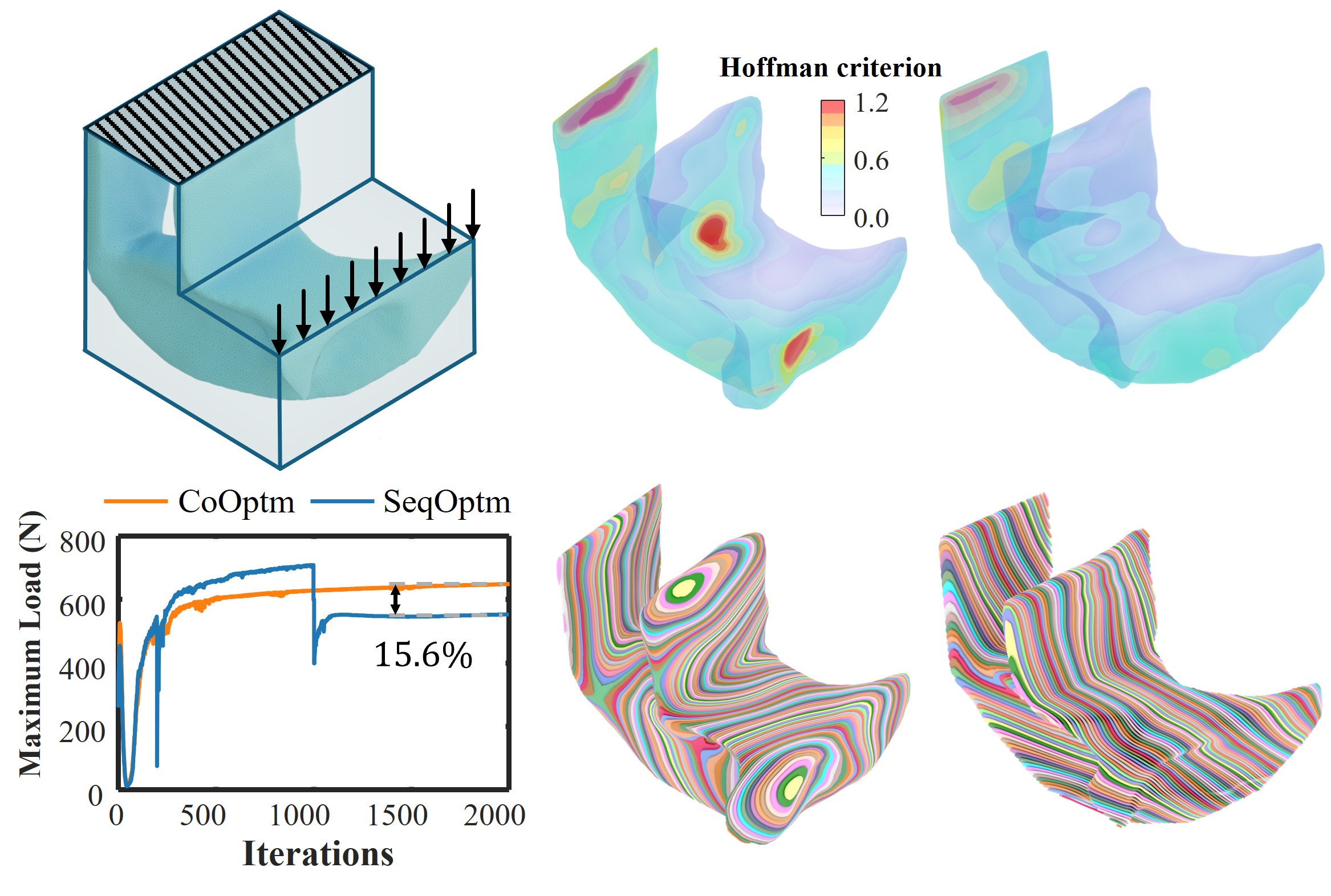}\vspace{-8pt}
\put(-240,80){\footnotesize \color{black}(a)}
\put(-140,80){\footnotesize \color{black}(c.1) Seq. Optm. I + II}
\put(-70,80){\footnotesize \color{black}(c.2) Our method}
\put(-240,2){\footnotesize \color{black}(b)}
\put(-140,2){\footnotesize \color{black}(d.1) Seq. Optm. I + II}
\put(-70,2){\footnotesize \color{black}(d.1) Our method}
\put(-88,92){\footnotesize \color{black}$F= 0.656~\text{kN}$}
\put(-240,155){\footnotesize \color{black}$V_f=25\%$}
\caption{Comparison of sequential optimization vs. our co-optimization for the L-Bracket model: (a) design domain; (b) evolution of maximum load during the optimization iterations; (c) the Hoffman criterion distribution; (d) the resultant curved layers for 5-axis 3D printing. 
}\label{fig:Lbracket}
\end{figure}

The second example is the L-Bracket as shown in Fig.~\ref{fig:Lbracket}(a), where its design domain is enclosed within a cubic region as $150 \times 150 \times 150 \, \text{mm}^3$. A distributed load is applied along the right edge of the design domain. The maximum yield-free load of co-optimization is 15.6\% higher than the result of sequential optimization (see Fig.~\ref{fig:Lbracket}(b)). Again, large yielded area can be found from the failure index distribution by Hoffman criterion on the result of sequential optimization as shown in Fig.~\ref{fig:Lbracket}(c), which is computed by the commercial FEA software. No yield occurs on the result of co-optimization. Their corresponding curved layers are as given in Fig.~\ref{fig:Lbracket}(d).

\begin{figure}
\centering
\includegraphics[width=\linewidth]{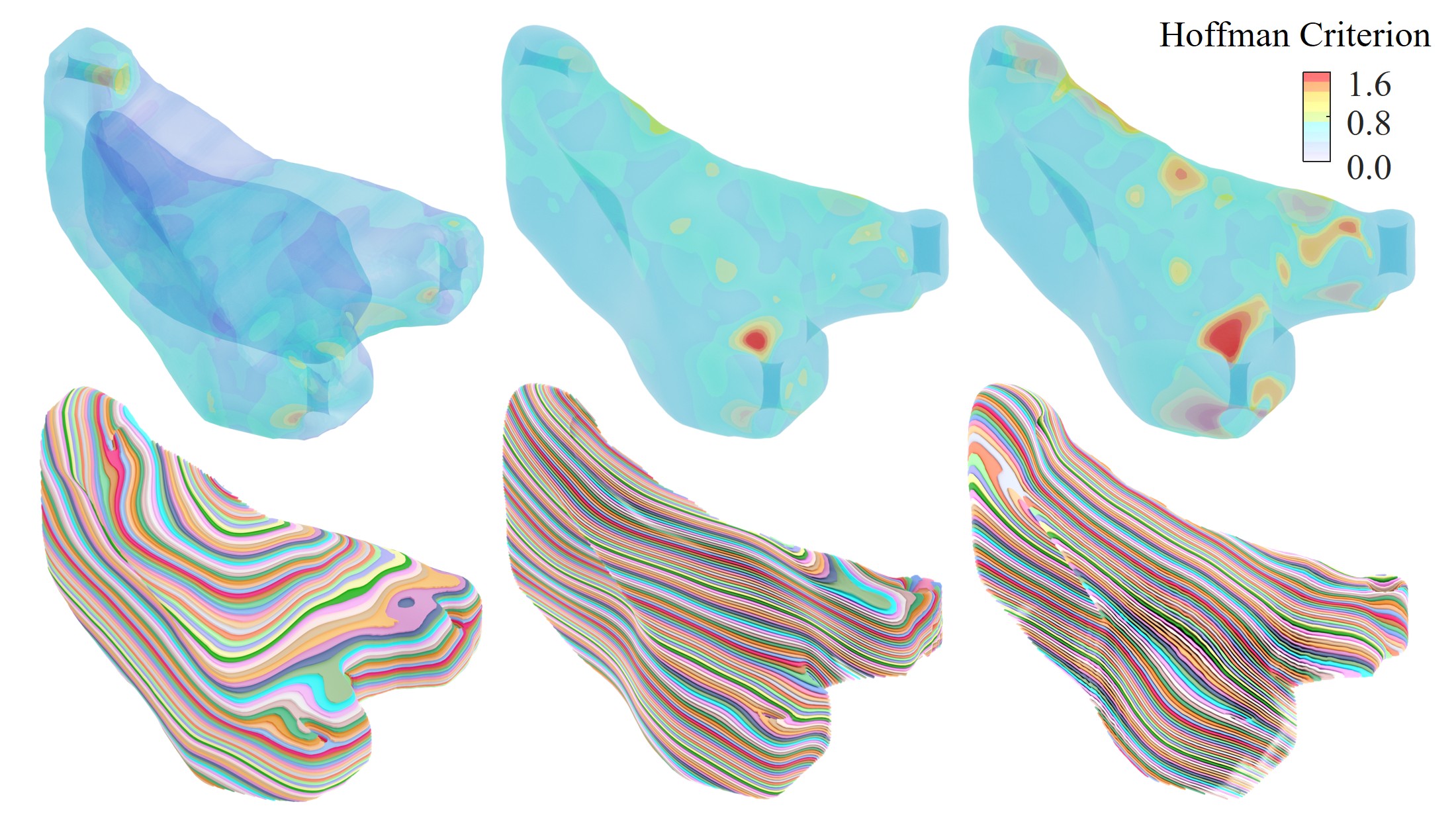}
\put(-235,5){\footnotesize \color{black}(a)}
\put(-155,5){\footnotesize \color{black}(b)}
\put(-75,5){\footnotesize \color{black}(c)}
\put(-210,125){\footnotesize \color{black}$F=1.923~\text{kN}$}
\caption{Comparison of different optimization strategies: (a) our method, (b) Seq. Optm. I + II, and (c) Seq. Optm. I + Neural Slicer~\cite{Liu2024NeuralSlicer}. 
}\label{fig:ABracketNNSclier}
\end{figure}
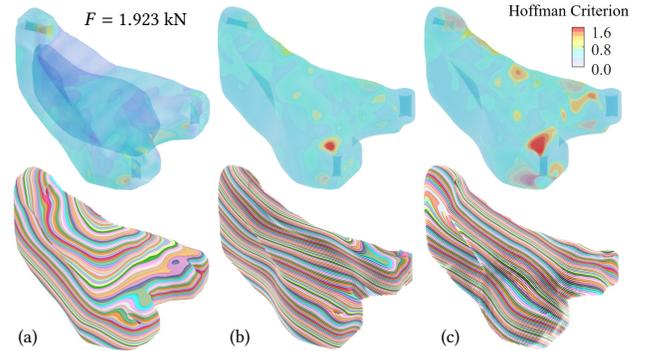

\subsubsection{Sequential optimization with fixed fiber-orientations}\label{subsubSec:SeqOptm_NeuralSlicer} 
To further demonstrate the advantages of our method, we conducted an additional study on the A-Bracket example for comparing ours with the result obtained by using Neural Slicer~\cite{Liu2024NeuralSlicer} to generate curved layers. In this comparison, the fiber orientations from Phase I of sequential optimization were provided as input to Neural Slicer. The resultant curved layers are shown in Fig. \ref{fig:ABracketNNSclier}(c). After that, the fiber orientations were projected onto the curved layers to generate toolpaths and form the material frames for anisotropic FEA. This result was then compared with those from our method and the sequential optimization, as illustrated in Fig.\ref{fig:ABracketNNSclier}(a) and (b), respectively. The failure index distributions were all evaluated under the maximum yield-free force of our method and are presented in Fig.\ref{fig:ABracketNNSclier}. It can be observed that our method achieves the best result.

It is worth noting that the Hoffman criterion is not explicitly optimized when generating curved layers using Neural Slicer. In contrast, the method used to generate curved layers in Seq. Optm. II simultaneously optimizes fiber orientations by adjusting $m(\cdot)$ and $a(\cdot)$ while maintaining the structural topology by fixing $\rho(\cdot)$. This explains why fewer yielded regions are observed in the result of Seq. Optm. I + II (see Fig. \ref{fig:ABracketNNSclier}(b)) when comparing to the result of Seq. Optm. I + Neural Slicer (see Fig. \ref{fig:ABracketNNSclier}(c)).

\begin{figure}
\centering
\includegraphics[width=\linewidth]{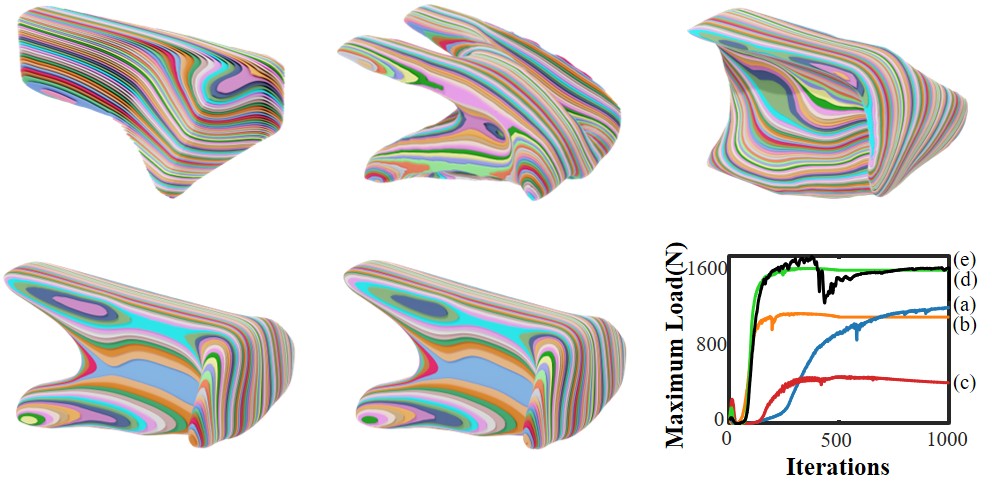}
\put(-242,66){\footnotesize \color{black}(a)~$1$ MLP-Layer}
\put(-213,120){\footnotesize \color{black}$F_{yd}=0.43\text{kN}$}
\put(-165,66){\footnotesize \color{black}(b)~$1$ MLP-Layer + F.E.}
\put(-125,120){\footnotesize \color{black}$F_{yd}=1.22\text{kN}$}
\put(-80,66){\footnotesize \color{black}(c)~$1$ RBF-Layer}
\put(-50,120){\footnotesize \color{black}$F_{yd}=1.12\text{kN}$}
\put(-242,5){\footnotesize \color{black}(d)~$5$ MLP-Layers}
\put(-213,55){\footnotesize \color{black}$F_{yd}=1.60\text{kN}$}
\put(-165,5){\footnotesize \color{black}(e)~$10$ MLP-Layers}
\put(-128,55){\footnotesize \color{black}$F_{yd}=1.62\text{kN}$}
\put(-80,5){\footnotesize \color{black}(f)}
\caption{The results of the cantilever beam example obtained by using different variations of neural network (a-e) and their convergence curves of computation (f). The maximum yield-free load $F_{yd}$ is also reported for different results.}
\label{fig:netAblation}
\end{figure}
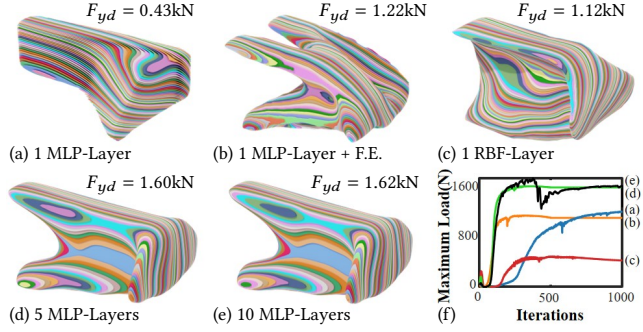

\subsubsection{Ablation study of network architecture}\label{subsubSec:networkAblation} 
In our framework, the design variables are the neural network coefficients. Reducing the number of neurons or adopting different types of implicit representations can lower the dimensionality of these design variables, potentially improving computational efficiency. We tested several variants of the network using the cantilever beam example (Fig.\ref{fig:overview}) to evaluate their performance of optimization as the maximum yield-free load $F_{yd}$. These include adopting different number of MLP layers, replacing MLP by the \textit{Radial Basis Functions} (RBF), and also using MLP layer with additional \textit{Fourier Encoding} (F.E.) (ref.~\cite{CHANDRASEKHAR2022Approximate}). The results are shown in Fig.\ref{fig:netAblation} with the convergence curves for different network variations. Among them, the current implementation using a five-layer MLP achieves the best trade-off between structural performance and computational efficiency.

\subsubsection{Resolution of FEA}\label{subsubSec:resolutionFEA} 
The resolution of FEA grids can influence the performance of optimized structures. To study this, we tested different grid resolutions in our framework while keeping other parameters unchanged using the cantilever beam example. The volumes of yielded regions ($V_{yd}$) under a load of $F=1.62~\text{kN}$ are evaluated using the FEA software Abaqus. We have
\begin{itemize}
\item Grid Res.~$45 \times 30 \times 30$: $V_{yd}=1.45\%$;
\item Grid Res.~$30 \times 20 \times 20$: $V_{yd}=1.72\%$;
\item Grid Res.~$15 \times 10 \times 10$: $V_{yd}=3.52\%$.
\end{itemize}
These results indicate that employing higher-resolution grids can slightly improve the performance of the resulting fiber-reinforced structures by reducing the volume of yielded regions; however, high-resolution will reduce the computational efficiency.

\subsubsection{Fiber orientation and toolpath consistency}
In addition to the grid resolution for FEA, we also examine the resolution of MC algorithm for curved layer and toolpath generation. The evaluation was conducted on a representative layer (shown on the far right in Fig. \ref{fig:overview}) by measuring the average angular error ($E_{avg}$) and its standard deviation ($E_{std}$) between the optimized fiber orientations (i.e., the cross-product of field gradients) and the toolpaths generated by \cite{zhang2024toolpath}. The results are as follows.
\begin{itemize}
\item MC Res.~$512^3$: $E_{avg}=3.32^\circ$ and $E_{std}=1.21^\circ$;
\item MC Res.~$256^3$: $E_{avg}=4.41^\circ$ and $E_{std}=1.47^\circ$;
\item MC Res.~$128^3$: $E_{avg}=7.57^\circ$ and $E_{std}=3.35^\circ$.
\end{itemize}
Using a higher resolution can slightly improve the alignment between the optimized fiber orientations and the toolpaths, although the improvement is not substantial.

\subsection{Manufacturability}
A key contribution of our co-optimization framework is its ability to ensure the manufacturability of the resulting fiber-reinforced composites. This important feature is demonstrated and validated through the examples presented in this sub-section.

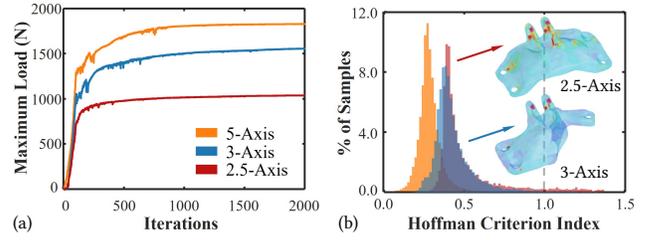
\begin{figure}
\centering
\includegraphics[width=\linewidth]{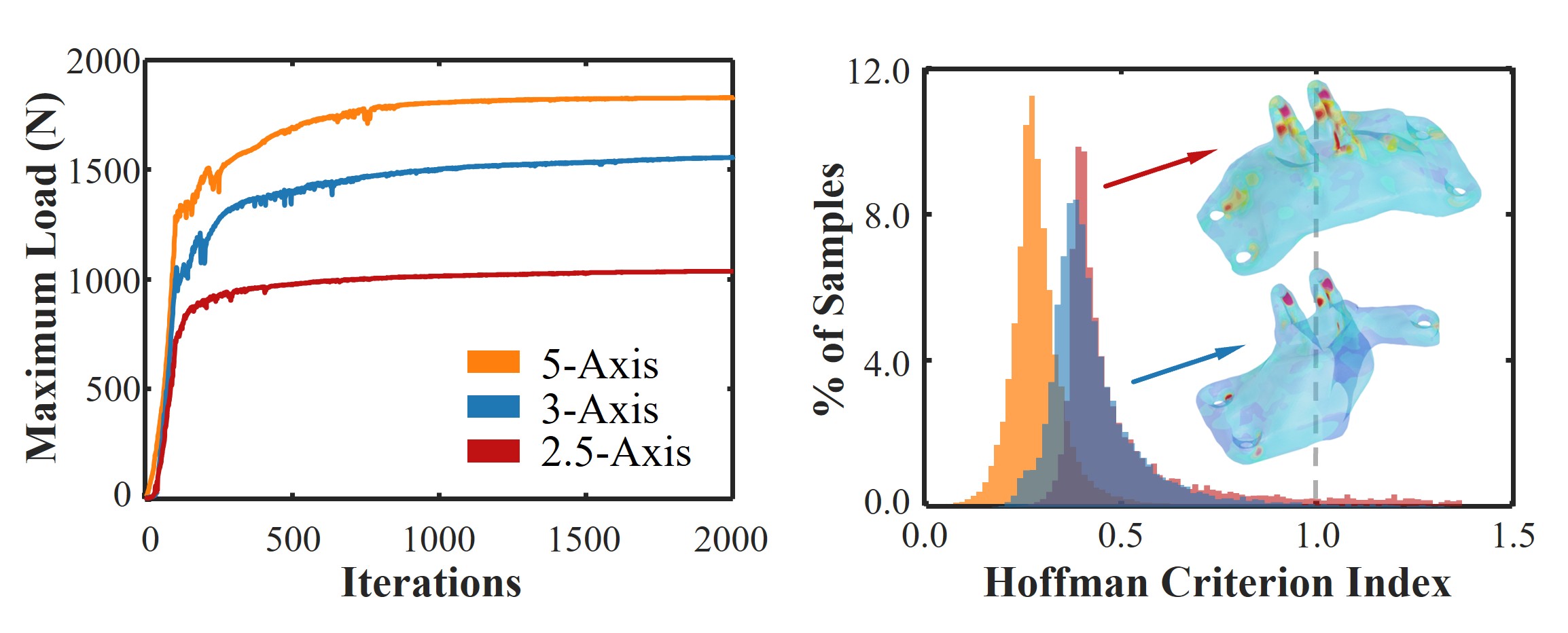}
\put(-33,58){\footnotesize \color{black} 2.5-Axis}
\put(-33,25){\footnotesize \color{black} 3-Axis}
\put(-240,6){\footnotesize \color{black} (a)}
\put(-117,6){\footnotesize \color{black} (b)}
\caption{The results of GE-Bracket by allowing different DoFs in motion: (a) the convergence curves of maximum yield-free loads, and (b) the failure index distributions w.r.t. Hoffman criterion when applying the force $F=1.855~\text{kN}$ -- i.e., the maximum yield-free force of 5-axis based optimization. 
}\label{fig:GEBracket_DiffMotionHoffman}
\end{figure}

\subsubsection{Different DoFs in motion}
We have tested our framework for solving the GE-Bracket and A-Bracket problems by allowing different DoFs in motion for different 3D printing hardware -- i.e., 5-axis, 3-axis, and 2.5-axis. Here half-axis denotes the axis that can only move in an asynchronous way. Limited by the function of controller, many desktop-level 3D printers can only conduct 2.5-axis motion to fabricate planar layers. 

The resultant structures, layers and toolpaths for the GE-Bracket example have been shown in Fig.\ref{fig:teaser} together with their corresponding motions illustrated. The convergence curves of optimization are given in Fig.\ref{fig:GEBracket_DiffMotionHoffman}(a). The failure index distributions in terms of Hoffman criterion are displayed in Fig.\ref{fig:GEBracket_DiffMotionHoffman}(b) together with their convergence curves. The 3-axis result is particularly interesting, as the optimizer has completely removed the material around one of the bolt holes. This outcome is primarily driven by the challenges of manufacturability. Without applying manufacturing constraints (by setting 
$\omega_{lc}$, $\omega_{ort}$ and $\omega_{lt}$ in Eq.\eqref{eq:TotalLoss_3x} to \textit{zero}), the 3-axis optimization would be the same result as `Seq. Optm. I' in Fig.\ref{fig:teaser}. However, by incorporating manufacturing constraints, the optimizer is forced to compromise, resulting in the omission of material around one bolt hole. Consequently, the maximum yield-free force is reduced to $1.531~\text{kN}$ and $1.035~\text{kN}$ for 3-axis and 2.5-axis respectively.

\begin{figure}
\centering
\includegraphics[width=\linewidth]{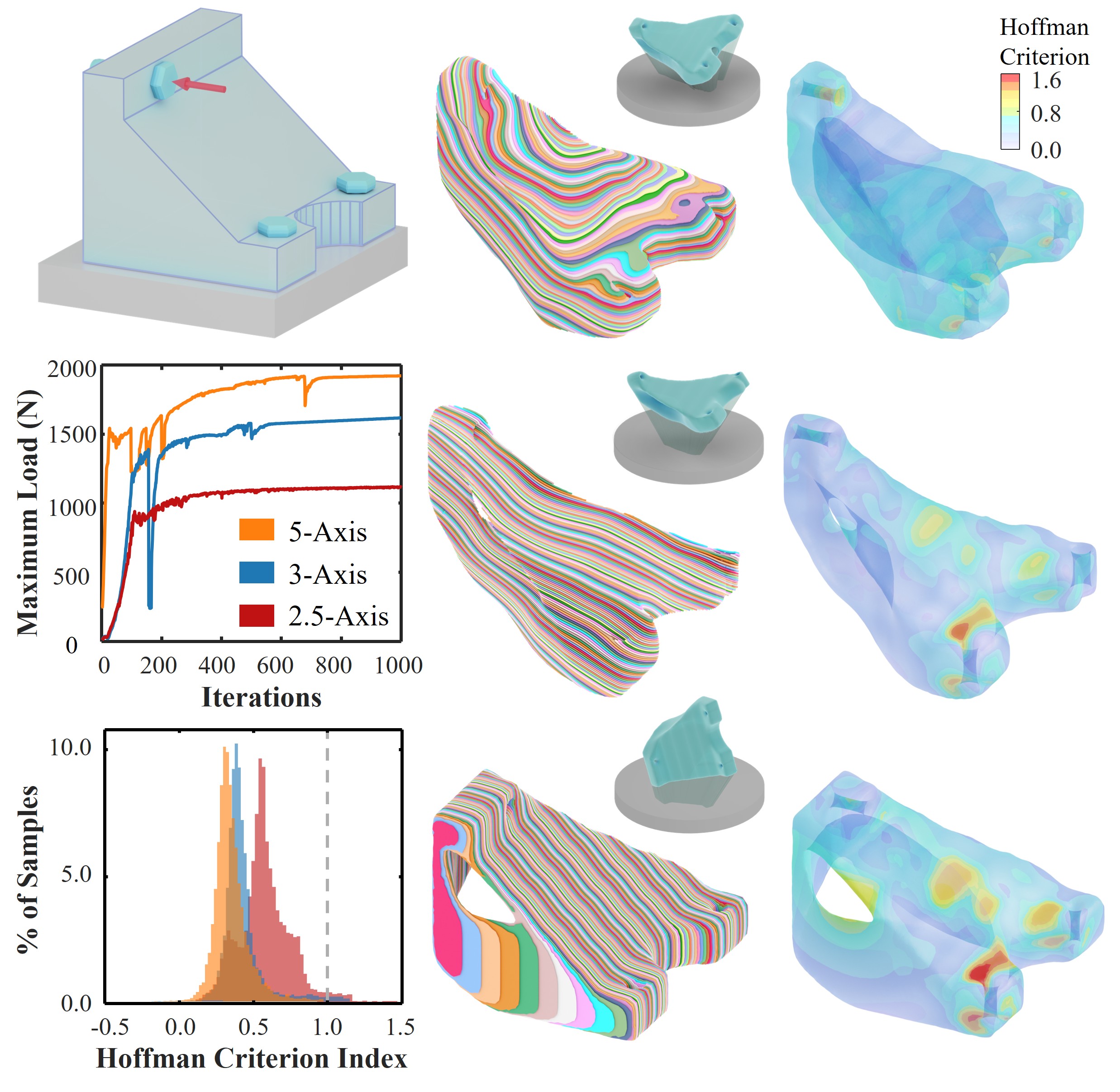}
\put(-240,163){\footnotesize \color{black}(a)}
\put(-145,163){\footnotesize \color{black}(b.1)}
\put(-70,163){\footnotesize \color{black}(b.2)}
\put(-240,88){\footnotesize \color{black}(c)}
\put(-145,88){\footnotesize \color{black}(d.1)}
\put(-70,88){\footnotesize \color{black}(d.2)}
\put(-240,5){\footnotesize \color{black}(e)}
\put(-145,5){\footnotesize \color{black}(f.1)}
\put(-70,5){\footnotesize \color{black}(f.2)}
\put(-140,230){\footnotesize \color{black}Pnt. Setup}
\put(-140,152){\footnotesize \color{black}Pnt. Setup}
\put(-140,73){\footnotesize \color{black}Pnt. Setup}
\put(-240,230){\footnotesize \color{black}$V_f=15\%$}
\caption{Comparison of A-Bracket optimization results by allowing motions in different DoFs: (a) design domain and boundary conditions, (b.1) the layers and the printing setup for 5-axis printing, (b.2) the failure index distribution of Hoffman criterion by applying the force $F=1.923~\text{kN}$, (c) convergence curves of maximum yield-free force during optimization, (d.1 \& d.2) the results for 3-axis printing, (e) the histogram comparison of failure indices, and (f.1 \& f.2) the results for planar-layer based printing (i.e., 2.5-axis). 
}\label{fig:ABracketL2}
\end{figure}
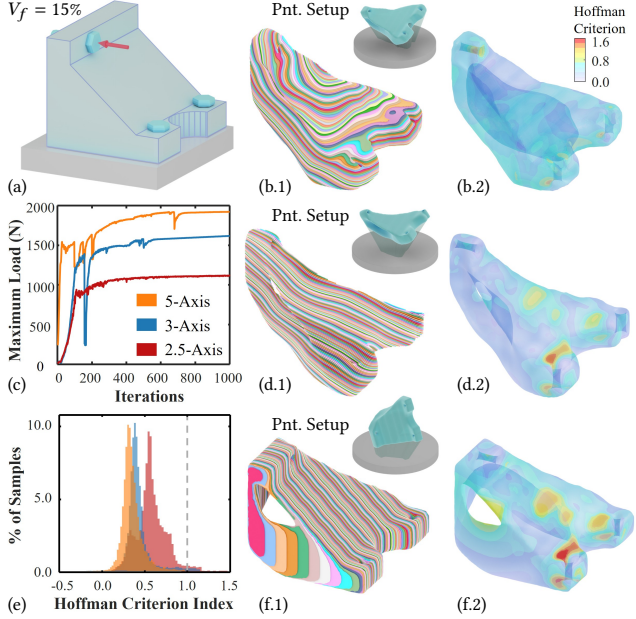

A similar comparison is given for the A-Bracket example as shown in Fig.\ref{fig:ABracketL2}, where all results are computed by using a target volume fraction as $15\%$. The layers that can be manufactured by motions in different DoFs are given in the middle column of Fig.\ref{fig:ABracketL2} together the setup orientations. The corresponding failure index distributions are presented in the right column of Fig.\ref{fig:ABracketL2}, which are generated by applying the force $F=1.923~\text{kN}$ (i.e., the maximum yield-free force of 5-axis result). The histograms of these distributions are also visualized in Fig.\ref{fig:ABracketL2}(e). In short, the strength of resultant composites can be reduced up to $41.9\%$ when imposing more constraints on the motion DoFs.

\begin{figure}
\centering
\includegraphics[width=\linewidth]{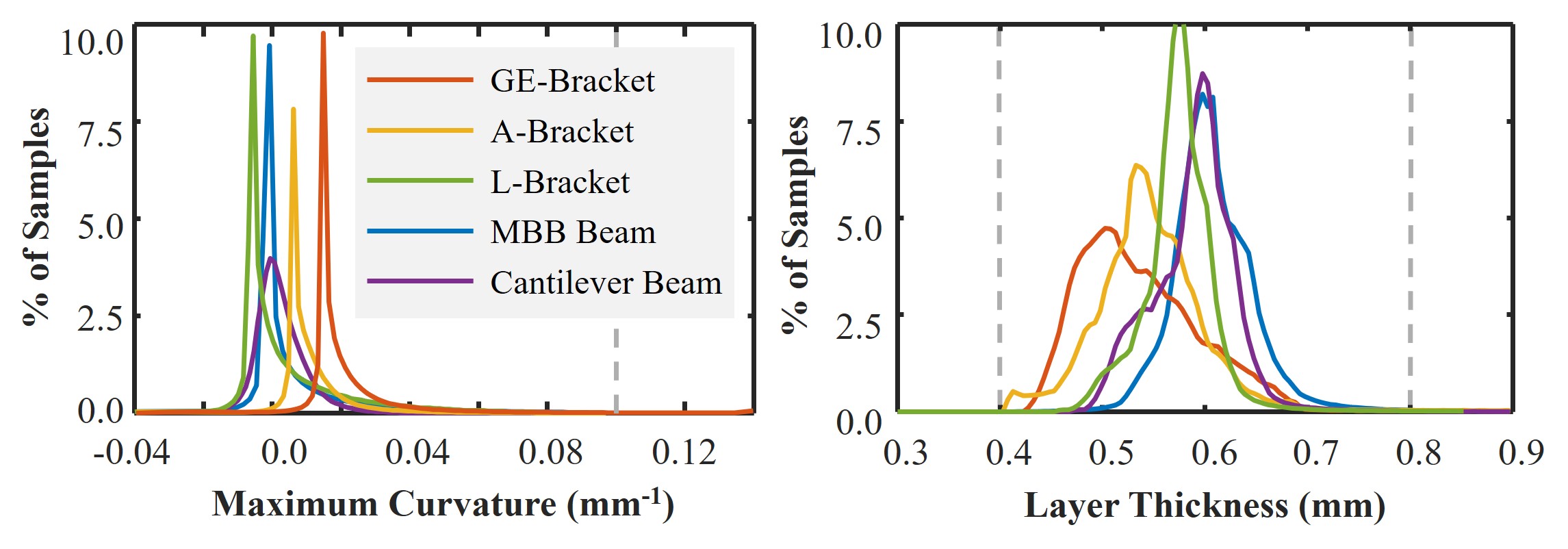}
\caption{Histograms of the maximum curvatures and the thickness on curved layers for optimized composites generated by our framework. Note that only sample points inside the resultant solids are employed for computing these histograms.
}\label{fig:CurvatureStastics}
\end{figure}

\begin{figure}
\centering
\includegraphics[width=\linewidth]{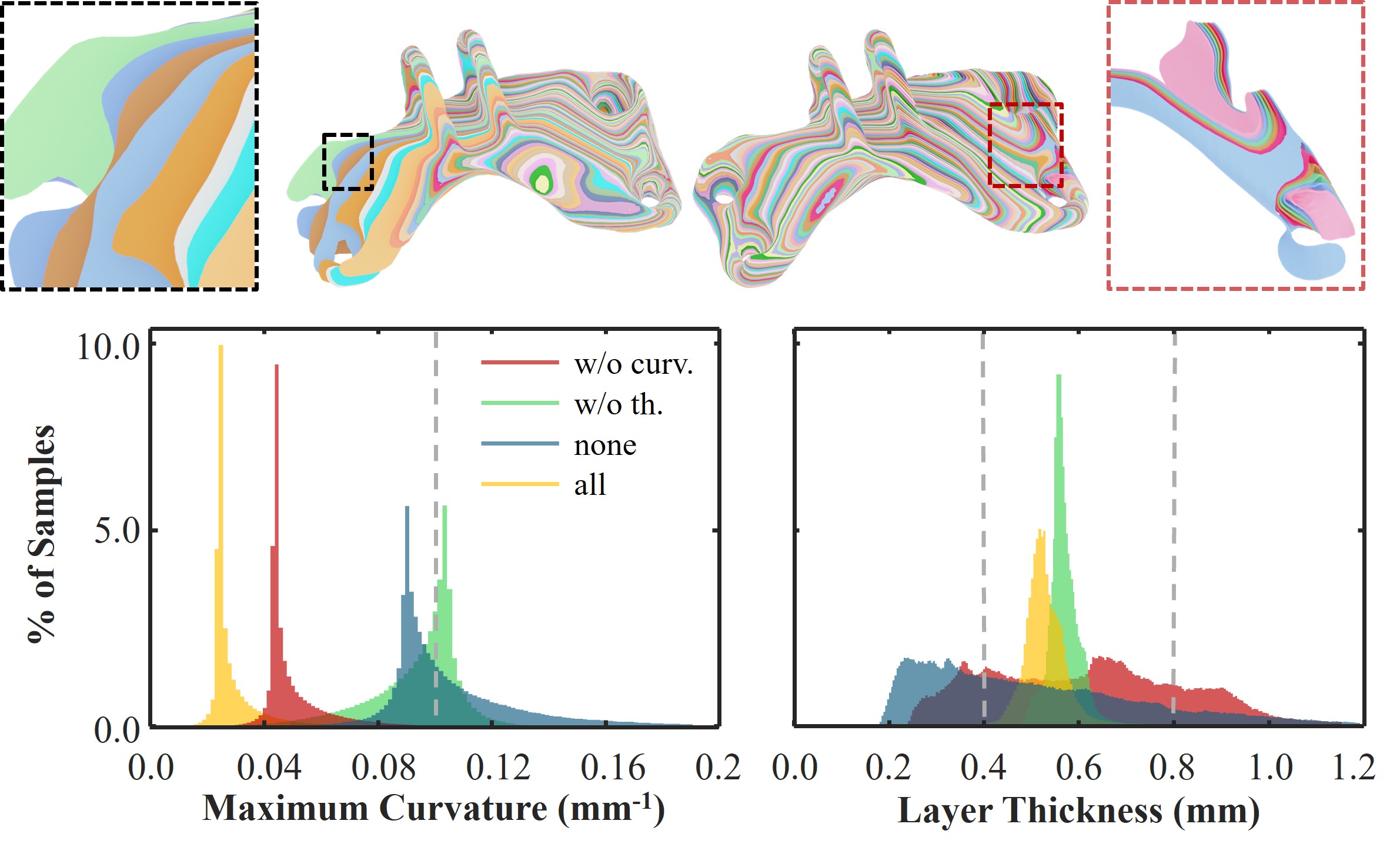}
\put(-165,100){\small \color{black}w/o th.}
\put(-95,100){\small \color{black}w/o curv.}
\caption{Ablation study for the maximum curvatures and the thicknesses on layers. The result of `none' is the same as `Seq. Optm. I' shown in Fig.\ref{fig:teaser} by omitting all terms of manufacturability. The `w/o curv.' result is generated by letting $\omega_{lc}=\omega_{mo}=0$, and `w/t th.' is obtained by setting $\omega_{lt}=0$. They are compared with the result by using `all' loss terms. Zoom-views present the problematic layers with large thickness variation (left) and maximum curvature (right). Again, only samples inside the structural solids are considered.
}\label{fig:AblationStudy}
\end{figure}

\subsubsection{Ablation study for layer thickness and curvature}
The layers generated by our method have their maximum curvature and layer thickness well controlled. As can be observed from Fig.\ref{fig:CurvatureStastics}, the results of all examples have their maximum curvatures lower than the allowed value $K_{lc}=0.1~\text{mm}^{-1}$ imposed in the local collision loss $\mathcal{L}_{lc}$ (i.e., Eq.~\eqref{eq:lossForCurvatureLC}). This can effectively avoid local collisions between the manufactured layers and the printer head. The variation of layer thickness is completely controlled within the range of $[0.4,0.8]~\text{mm}$ as well. 

The ablation study is then conducted for the losses of curvature and layer thickness. As shown in Fig.\ref{fig:AblationStudy}, layers with high curvature are generated by turning off the curvature control (i.e., letting $\omega_{lc}=\omega_{mo}=0$). Similarly, when turning off the layer thickness control by assigning $\omega_{lt}=0$, the results with large thickness variation are generated. The histograms in Fig.\ref{fig:AblationStudy} give the distributions of maximum curvature and layer thickness for this ablation study, which prove the effectiveness of these terms in our optimization framework. 

\begin{figure}
\centering
\includegraphics[width=\linewidth]{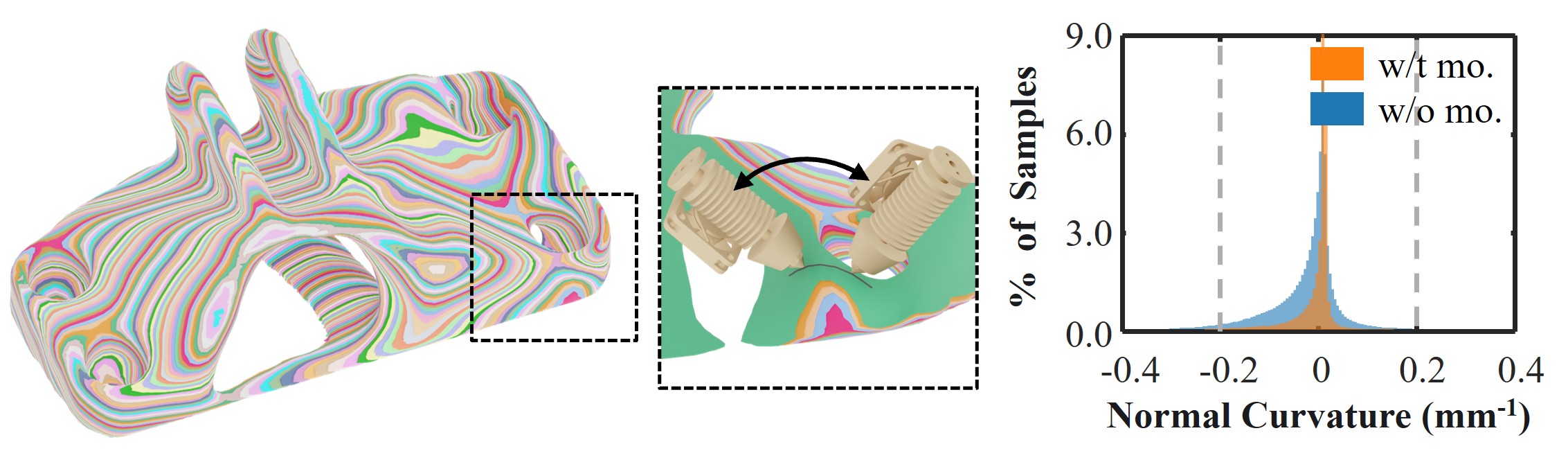}
\put(-190,5){\footnotesize \color{black}w/o mo.}
\caption{Ablation study on the path curvature. Without the loss $\mathcal{L}_{mo}$, the path curvature exceeds the specified bound of $K_{\max}^f = 0.2$, as evidenced by the problematic layer shown in the left and the histogram comparison of path curvatures in the right.}\label{fig:AblationStudyMotion}
\end{figure}

The inclusion of a path curvature loss, $\mathcal{L}_{mo}$, helps to prevent surface layers with large curvature along the toolpath direction. An ablation study was conducted to illustrate the results without $\mathcal{L}_{mo}$, as shown in the left of Fig.\ref{fig:AblationStudyMotion}. The histogram comparison of path curvatures, presented in the right of Fig.\ref{fig:AblationStudyMotion}, proves that the path curvatures can be effectively controlled within the range of $\pm 0.2$ by the loss $\mathcal{L}_{mo}$.

\subsubsection{Effectiveness of setup orientation loss}
\begin{figure}
\centering
\includegraphics[width=\linewidth]{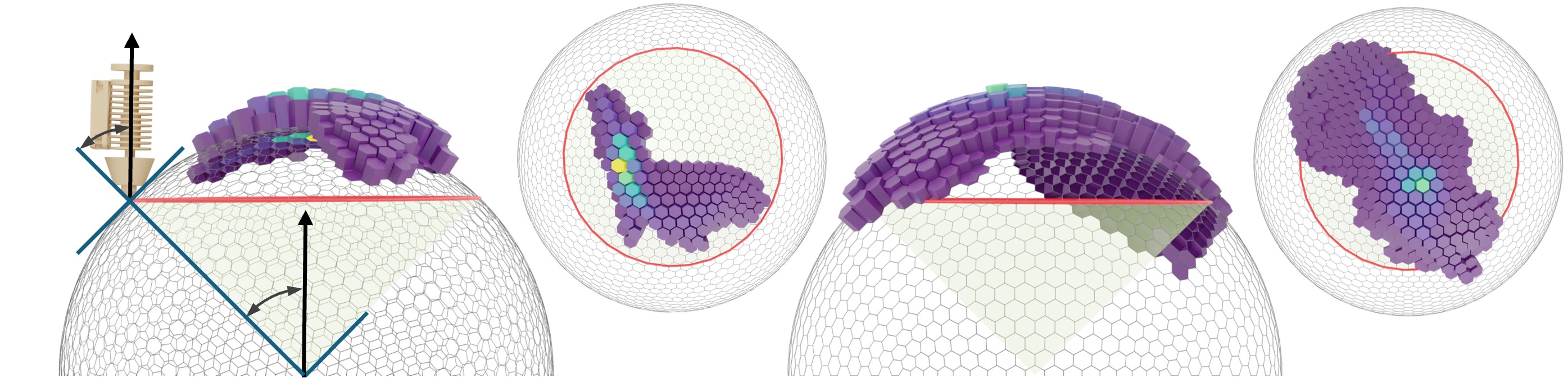}
\put(-200,48){\footnotesize \color{black}w/t ori.}
\put(-93,48){\footnotesize \color{black}w/o ori.}
\put(-193,20){\footnotesize \color{black}$\mathbf{n}$}
\put(-205,16){\footnotesize \color{black}$\beta$}
\put(-220,53){\footnotesize \color{black}$\mathbf{n}$}
\put(-236,42){\footnotesize \color{black}$\beta$}
\caption{Ablation study of the orientation loss. The histogram of surface normals on all layers is mapped onto the Gaussian sphere, highlighting the collision-free layers for a 3-axis machine should have all normals within the red circle. Results with vs. without the orientation loss $\mathcal{L}_{ort}$ are given. 
}\label{fig:AblationStudyOrt}
\end{figure}
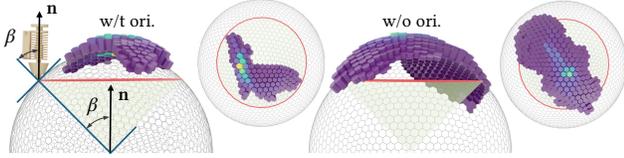

Different from 5-axis 3D printing, our co-optimization for 3-axis machine incorporates a setup orientation loss, $\mathcal{L}_{ort}$, to prevent local collisions. The effectiveness of $\mathcal{L}_{ort}$ is demonstrated in Fig.~\ref{fig:AblationStudyOrt}. The analysis is conducted by mapping the surface normals of all resultant curved layers onto the Gaussian sphere, where the collision-free region is highlighted within the red circle. A comparison of the results clearly shows that, without $\mathcal{L}_{ort}$, the surface normals are poorly controlled. Consequently, regardless of how the Gaussian sphere is rotated (i.e., by adjusting the setup orientation of the GE-Bracket), it becomes impossible to align all layers to let their surface normals remain within the collision-free red circle.

\subsection{Physical Experiments}
We conducted physical experiments to validate the manufacturability and the mechanical strength of composites generated by our co-optimization pipeline. The multi-axis motions with varying DoFs were realized on a robotic 3D printing system, consisting of an ABB IRB 2600 robotic arm with 6 DoFs and an ABBA 250 positioner with 2 DoFs. Using a dual print-head design, both support materials and part materials were printed continuously along different directions in the model's space. The models were fabricated using PLA-CF, while the supports were printed with water-soluble \textit{Polyvinyl Alcohol} (PVA). Motion planning and inverse kinematics computations were based on the publicly accessible code from \cite{zhang2021singular, Chen2025CoOptimization}.

\begin{figure}
\centering
\includegraphics[width=\linewidth]{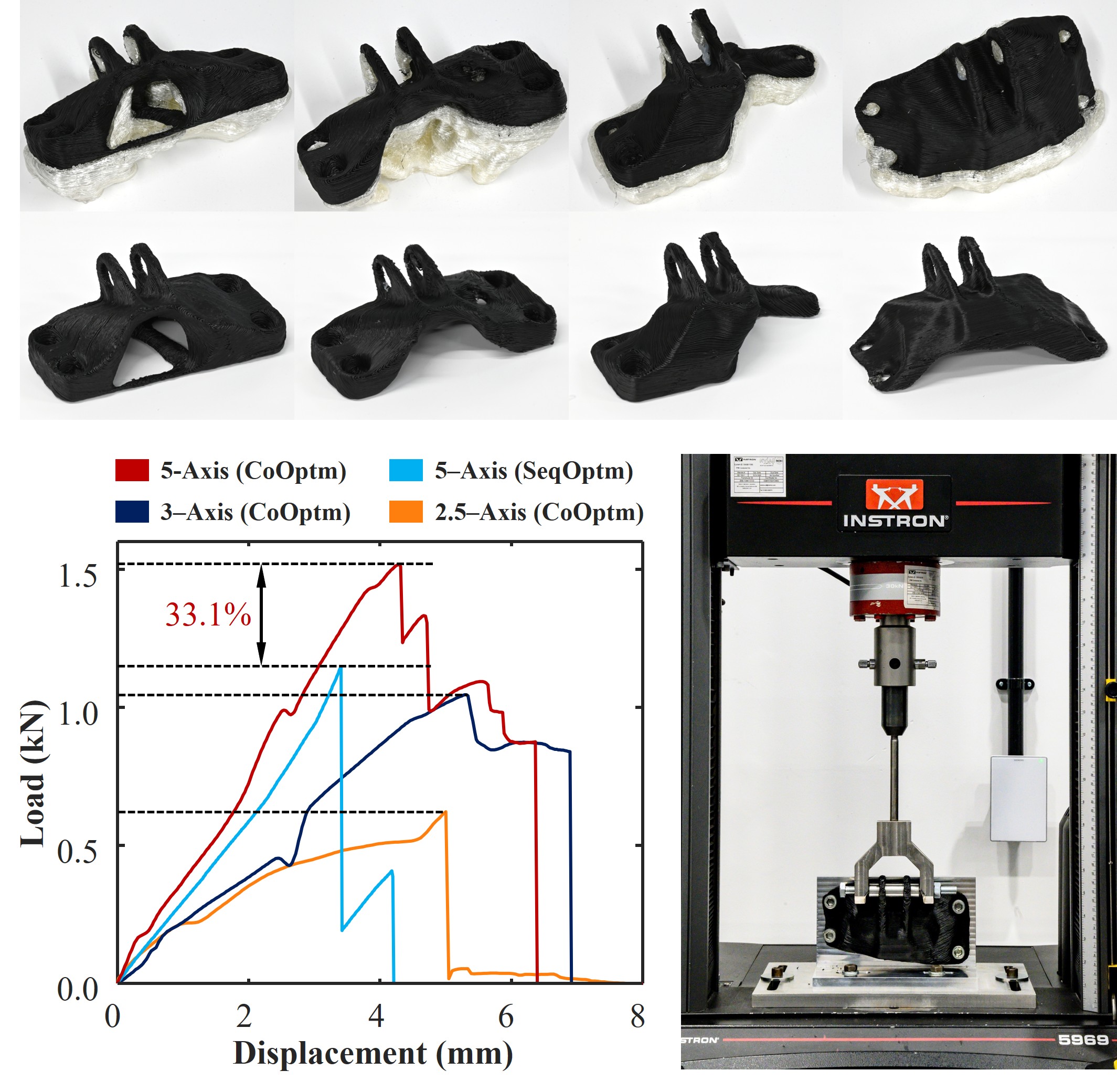}
\put(-205,192){\footnotesize \color{black}5-Axis}
\put(-205,185){\footnotesize \color{black}(CoOptm)}
\put(-145,192){\footnotesize \color{black}5-Axis}
\put(-145,185){\footnotesize \color{black}(SeqOptm)}
\put(-85,192){\footnotesize \color{black}3-Axis}
\put(-30,192){\footnotesize \color{black}2.5-Axis}
\put(-148,112){\footnotesize \color{black}$F_f=1.519~\text{kN}$}
\put(-148,91){\footnotesize \color{black}$F_f=1.141~\text{kN}$}
\put(-148,78){\footnotesize \color{black}$F_f=1.046~\text{kN}$}
\put(-148,65){\footnotesize \color{black}$F_f=0.622~\text{kN}$}
\caption{Physically prints of the optimized GE-Bracket composites using different DoFs in motion are shown together with their force-displacement curves obtained from tensile tests. 
}\label{fig:PhysicalResults}
\end{figure}

The results of physical fabrication, with and without supports, are shown at the top of Fig. \ref{fig:PhysicalResults}. Tensile tests of these specimens were conducted using an INSTRON 5960 tensile machine. The resultant force-displacement curves, along with the failure loads ($F_f$) for every cases, are presented in Fig. \ref{fig:PhysicalResults}. It can be observed that our co-optimization method produced a structure with a failure load that is $33.1\%$ higher than the structure generated by sequential optimization. 

\section{Discussion}\label{sec:Discussion}

\begin{figure*}
\centering
\includegraphics[width=\linewidth]{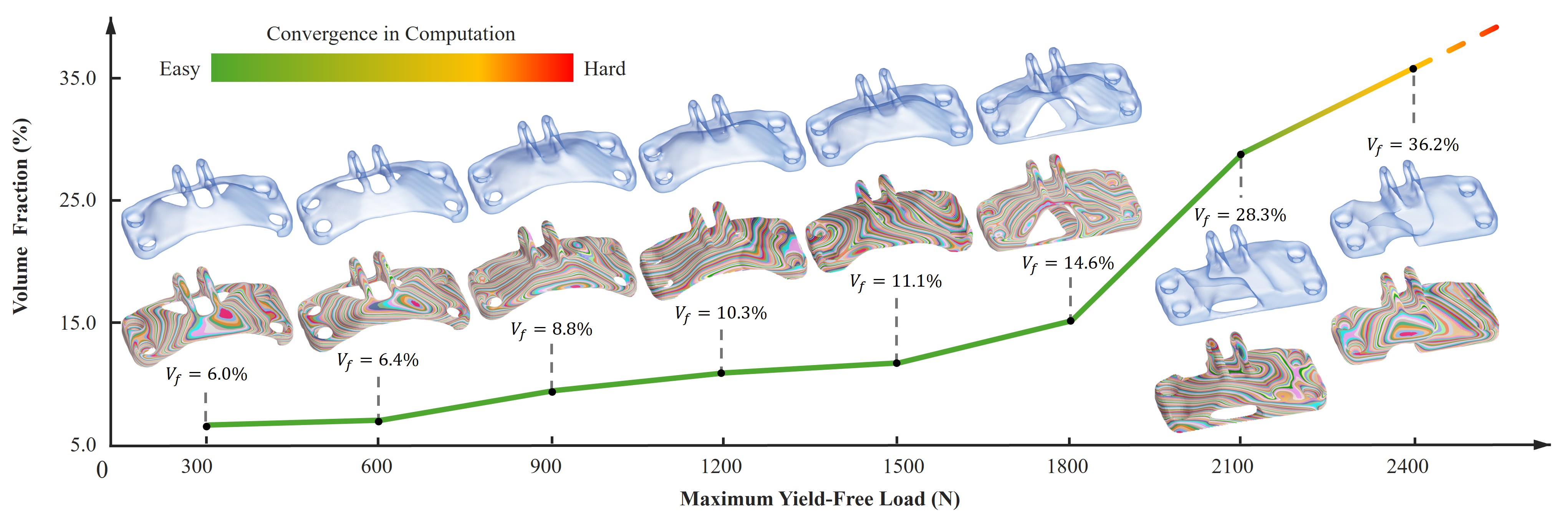}
\caption{The results of lightweight optimization by giving different maximum yield-free loads -- i.e., the generalization of our approach as defined in Eq.\eqref{eq:ObjFuncVol}.}
\label{fig:forceBasedMinVolOptm}
\end{figure*}

\subsection{Generality of Our Pipeline}
Although our paper primarily focuses on fiber-reinforced composites where the key mechanical strength is evaluated based on material failure using the Hoffman criterion, our computational pipeline is a general framework for concurrent optimization of structural topology and manufacturable layers for multi-axis 3D printing. Two feasible variants of design objectives are discussed below. 

\subsubsection{Stiffness-based optimization} 
Stiffness-based optimization can be realized by changing the loss function of design objective to the compliance energy as defined in Eq.\eqref{eq:Compliance}. That is
\begin{equation} \label{eq:LossStiffness}
    \mathcal{L}_{obj} := \Gamma(\theta_{\rho},\theta_m,\theta_a)=\mathbf{U}^T \mathbf{K}_{\rho,m,a} \mathbf{U} = \mathbf{U}^T \mathbf{F},
\end{equation}
where $\mathbf{K}_{\rho,m,a}$ is parameterized by the network coefficients $\boldsymbol{\theta}_{\rho}$, $\boldsymbol{\theta}_m$ and $\boldsymbol{\theta}_a$. The differentiation of this loss can be derived as
\begin{equation}
    \frac{\partial \mathcal{L}_{obj}}{\partial \theta} = -\frac{\partial \mathbf{U}^T\mathbf{K}\mathbf{U}}{\partial \theta}
        = - 2\mathbf{K}\mathbf{U}\frac{\partial \mathbf{U}}{\partial \theta} - \mathbf{U}\mathbf{U}^T\frac{\partial \mathbf{K}}{\partial \theta}. 
\end{equation}
When using the same total loss as defined in Eq.\eqref{eq:TotalLoss} and the same weighting scheme as our strength-based optimization, an example result of stiffness-based optimization can be found in Fig.\ref{fig:resMBBStrengthStiffness}. Compared with the result of strength-based optimization, using the stiffness loss as Eq.\eqref{eq:LossStiffness} leads to structures with small strains but easier to yield as indicated by the strain energy distribution and the failure index map. 

\subsubsection{Lightweight optimization} 
When designing a lightweight structure, instead of specifying the maximum allowed volume $V^*$, users often wish to compute the structure with minimal volume that satisfies the Hoffman criterion. This can be formulated as:
\begin{eqnarray}
            & \min &  \int_{\Omega} H(\rho(\mathbf{x})) \, d \mathbf{x}  \label{eq:ObjFuncVol} \\
         & s.t.& \mathbf{K}_{\rho,m,a} \mathbf{U} = \mathbf{F}, \nonumber \\
         & & \Gamma(\sigma_e) \leq 1, \quad \forall e \in \Omega, \nonumber \\
         && \Pi_j(\theta_\rho,\theta_m,\theta_a), \quad j=1,2,...,M. \nonumber
\end{eqnarray}
The problem can be solved by our computational pipeline by defining a volume based design loss term:
\begin{center}
    $\mathcal{L}_{obj} := \int_{\Omega} H(\rho(\mathbf{x})) \, d \mathbf{x}$
\end{center}
together with a loss for yield constraints as
\begin{center}
    $\mathcal{L}_{yd} :=  \sum_e \text{ReLU}(\Gamma(\boldsymbol{\sigma}_e)-1)$.
\end{center}
The total loss will be changed to include $\mathcal{L}_{yd}$ as 
\begin{center}
    $\mathcal{L}_{total} :=   \omega_{obj}\mathcal{L}_{obj} +  \omega_{yd}\mathcal{L}_{yd}+ \omega_{lc}\mathcal{L}_{lc} + \omega_{mo}\mathcal{L}_{mo} + \omega_{lt}\mathcal{L}_{lt}$
\end{center} 
by using the same weighting scheme as introduced in Sec.~\ref{subsec:WeightingScheme}. Results of lightweight designs for the GE-Bracket challenge by giving different maximum yield-free loads are shown in Fig.\ref{fig:forceBasedMinVolOptm}.

\subsection{Extension for Shape Control}
Our framework naturally supports the integration of shape control within the optimization process. For example, symmetry can be enforced either by directly constraining the density field (e.g., $\rho(x,y,z) = \rho(-x,y,z)$) or by symmetrizing it during finite element analysis (FEA) using a projection such as
\begin{center}
$\rho(x,y,z) = \max\{\rho(-x,y,z), \rho(x,y,z)\}$.
\end{center} 
The $\max(\cdot)$ function can be smoothly approximated using a $p$-norm formulation, enabling gradient-based optimization. Nevertheless, incorporating such shape control may significantly influence the final performance of the optimized structures.

\subsection{Limitation} 
The results of our approach, demonstrated through both computational experiments and physical validations, are highly promising. However, there are several aspects that need improvement in future work. First, the FEA computation in our current implementation is very time-consuming. The optimizer spends over 90\% of each iteration performing FEA, and this computational cost increases significantly with higher resolution requirements. To address this bottleneck, future research will explore leveraging PINNs and GPU-based acceleration to enhance computational efficiency.

The convergence and robustness of our framework are primarily influenced by improperly defined constraints during optimization. For instance, when an extremely small volume fraction is specified for TO, achieving a yield-free structure may become infeasible -- i.e., the total loss may fail to converge after a large number of iterations.

Our current optimization formulation is based on the yield metric using the Hoffman criterion. While the yield load can be readily identified for ductile materials, carbon fiber-reinforced composites exhibit predominantly brittle behavior, making the yield load difficult to measure in physical validation. As a result, although the mechanical strength in our optimization is formulated based on yield, physical validation relies on failure load measurements. In future work, we plan to incorporate advanced imaging techniques, such as \textit{computed tomography} (CT), to more accurately measure the yield behavior of composites. Additionally, our current formulation assumes linear material properties, which limits its applicability to handle materials with nonlinear elastic behavior.

Lastly, our framework adopts a conservative collision-free condition, where the allowed curvature, $K_{lc}$, is typically chosen as a value smaller than that of the bounding sphere of the printer nozzle. This conservative choice accounts for tolerances arising from hardware errors. Future work will aim to refine this condition while maintaining robustness in real-world fabrication.

\section{Conclusion}
In this paper, we have presented a manufacturability integrated design optimization framework for fiber-reinforced composites, where the structural topology, the manufacturable layers, and the fiber orientations are optimized simultaneously. 
Our method integrates three implicit neural fields to represent geometric shapes, layer sequences, and fiber orientations so that enables a unified and differentiable neural optimization pipeline. By embedding objectives such as anisotropic strength, structural volume, machine motion control, layer curvature, and layer thickness as loss functions, the composites generated by our framework can achieve superior mechanical performance while maintaining manufacturability for filament-based multi-axis 3D printing on hardware platforms with different DoFs in motion. Physical experiments have validated the manufacturability and the enhanced mechanical strength through tensile tests. To the best of our knowledge, this is the first approach that simultaneously optimizes the anisotropic strength of fiber-reinforced composites while ensuring their manufacturability directly.

\begin{acks}
    The project is supported by the chair professorship fund at the University of Manchester and UK Engineering and Physical Sciences Research Council (EPSRC) Fellowship Grant (Ref.\#: EP/X032213/1).
\end{acks}

\bibliographystyle{ACM-Reference-Format}
\bibliography{reference}

\appendix
\end{document}